\documentclass{article}

\PassOptionsToPackage{numbers, compress}{natbib}

\usepackage[preprint]{neurips_2024}

\usepackage[utf8]{inputenc} 
\usepackage[T1]{fontenc}    
\usepackage{hyperref}       
\usepackage{url}            
\usepackage{booktabs}       
\usepackage{amsfonts}       
\usepackage{nicefrac}       
\usepackage{microtype}      
\usepackage{xcolor}         

\usepackage{subcaption}
\usepackage{graphicx}

\usepackage{amsmath}
\usepackage{amssymb}
\usepackage{mathtools}
\usepackage{amsthm}
\usepackage{float}

\usepackage{algorithm}
\usepackage{algorithmic}
\usepackage{alphalph}

\usepackage{wrapfig}

\newtheorem{theorem}{Theorem}[section]
\newtheorem{proposition}[theorem]{Proposition}

\theoremstyle{definition}

\theoremstyle{remark}

\DeclareMathOperator*{\argmax}{arg\,max}
\DeclareMathOperator*{\argmin}{arg\,min}

\newcommand{\ie}{\textit{i}.\textit{e}.}
\newcommand{\eg}{\textit{e}.\textit{g}.}
\newcommand{\etal}{\textit{et} \textit{al}.}
\newcommand{\etc}{\textit{etc}.}

\title{Robust Multi-Agent Reinforcement Learning by Mutual Information Regularization}

\begin{document}

\maketitle

\begin{abstract}

In multi-agent reinforcement learning (MARL), ensuring robustness against unpredictable or worst-case actions by allies is crucial for real-world deployment. Existing robust MARL methods either approximate or enumerate all possible threat scenarios against worst-case adversaries, leading to computational intensity and reduced robustness. In contrast, human learning efficiently acquires robust behaviors in daily life without preparing for every possible threat. Inspired by this, we frame robust MARL as an inference problem, with worst-case robustness implicitly optimized under all threat scenarios via off-policy evaluation. Within this framework, we demonstrate that \underline{M}utual \underline{I}nformation \underline{R}egularization as \underline{R}obust \underline{R}egularization (MIR3) during routine training is guaranteed to maximize a lower bound on robustness, without the need for adversaries. Further insights show that MIR3 acts as an information bottleneck, preventing agents from over-reacting to others and aligning policies with robust action priors. In the presence of worst-case adversaries, our MIR3 significantly surpasses baseline methods in robustness and training efficiency while maintaining cooperative performance in StarCraft II and robot swarm control. When deploying the robot swarm control algorithm in the real world, our method also outperforms the best baseline by 14.29\%.

\end{abstract}


\section{Introduction}
Cooperative multi-agent reinforcement learning (MARL) \cite{lowe2017maddpg,rashid2018qmix,yu2021mappo,kuba2021hatrpo} has advanced in a wide variety of challenging scenarios, including StarCraft II \cite{vinyals2019alphastar}, Dota II \cite{berner2019openaifive}, \etc \ In real world, however, current MARL algorithms falls short when agents' actions deviate from their original policy, or facing adversaries performing worst-case actions \cite{li2019M3DDPG,sun2022romax,gleave2019iclr2020advpolicy,ly2021hijacking,dinh2023nonoblivious,li2023ami}. This greatly limits the potential of MARL in real world, notably in areas such as robot swarm control \cite{huttenrauch2019swarm,chen2022dexhand}.

Research on robust multi-agent reinforcement learning (MARL) against action uncertainties primarily focuses on max-min optimization against worst-case adversaries \cite{li2019M3DDPG,sun2022romax,zhang2020rmaddpg,nisioti2021romq,li2023epr-mappo}. This approach can be framed as a zero-sum game \cite{zhang2020rmaddpg,tessler2019actionrobustmannor}, where defenders with fixed parameters during deployment aim to maximize performance despite unknown proportions of adversaries employing worst-case, non-oblivious adversarial policies \cite{gleave2019iclr2020advpolicy,dinh2023nonoblivious}. However, in a multi-agent context, each agent can be perturbed, leading to an exponential increase in potential threat scenarios, making max-min optimization against each threat intractable. To address this complexity, some methods \cite{li2019M3DDPG,sun2022romax,bukharin2024ernie} approximate the problem by treating all agents as adversaries, resulting in overly pessimistic or ineffective policies. Others attempt to enumerate all threat scenarios \cite{nisioti2021romq,li2023epr-mappo,yuan2023evolution}, but often struggle to explore each threat scenario sufficiently during training, leaving defenders still vulnerable to worst-case adversaries. Consequently, max-min optimization provides limited defense capabilities in MARL and incurs high computational costs \cite{derman2021twiceregularized}.

\begin{wrapfigure}{r}{0.5\linewidth}
  \centering
  \includegraphics[width=\linewidth]{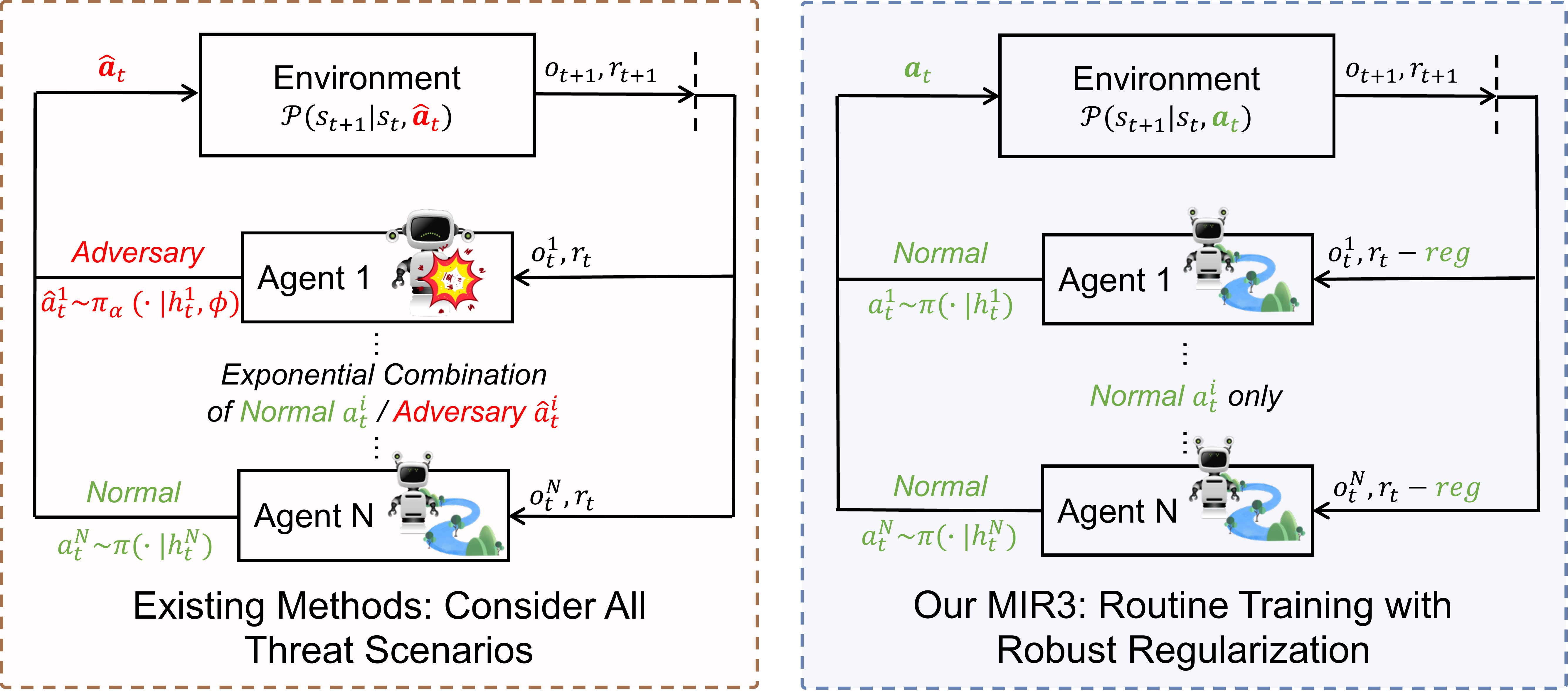}
  \caption{Our policies are learned under routine scenarios but are provably robust against unseen worst-case adversaries through robust regularization, contrasting with existing approaches that require exposure to all possible threat scenarios.}
  \label{intro}
  \vspace{-0.1in}
\end{wrapfigure}

Instead of explicitly considering every threat scenario, human learns through experiences in routine scenarios without an ``adversary'', but are able to react under diverse unseen threats. Motivated by this, we propose \underline{M}utual \underline{I}nformation \underline{R}egularization as \underline{R}obust \underline{R}egularization (MIR3) for robust MARL. As depicted in Fig. \ref{intro}, rather than requiring exposure to all threat scenes, our policies are learned in routine scenarios, but provably robust when encountering unseen worst-case adversaries. Specifically, we model this objective as an inference problem \cite{levine2018probinf}. Policies are designed to simultaneously maximize cooperative performance in an attack-free environment and ensure robust performance through off-policy evaluation \cite{levine2020offlinereview}. Within this framework, we proof that under specific conditions, regularizing the mutual information between histories and actions can maximize the robustness lower bound across all threat scenarios, without requiring specific adversaries.

Beyond theoretical derivations, MIR3 can be understood as an information bottleneck \cite{tishby2000ib} or as learning a task-relevant robust action prior \cite{grau2018actionprior}. From the information bottleneck perspective, our goal is to learn a policy that solves the task using minimum sufficient information of current history. Therefore, it suppresses false correlations in the policy created by action uncertainties and minimizes agents' overreactions to adversaries, fostering robust agent-wise interactions. From the standpoint of robust action prior, we aim to restrict the policy from deviating from a prior action distribution which is not only generally favored by the task, but also maintains intricate tactics under attack. Experiments in StarCraft II and rendezvous environments shows MIR3 demonstrates higher robustness against worst-case adversaries and requires less training time than max-min optimization approaches, on both QMIX and MADDPG backbones. When the magnitude of regularization are properly chosen, we find suppressing mutual information will not negatively affect cooperative performance, but even slightly enhance it. Finally, the superiority of MIR3 remains consistent when deployed in real world robot swarm control scenario, outperforming the best performing baseline by 14.29\%.

\paragraph{Contribution.} Our contributions are two-folded. First, inspired by human adaptability, we propose MIR3 that efficiently trains robust MARL policies against diverse threat scenarios without adversarial input. Second, we theoretically frame robust MARL as an inference problem and optimize robustness via off-policy evaluation. In this framework, we proof our MIR3 maximizes a lower bound of robustness, reducing spurious correlations and learning robust action prior. Empirically, experiments on six StarCraft II tasks and robot swarm control shows our MIR3 surpasses baselines in robustness and training efficiency, while maintaining cooperative performance on MADDPG and QMIX backbones. This superiority is consistent when deploying the algorithm in real world.


\section{Preliminaries}
\paragraph{Cooperative MARL as Dec-POMDP.} We formulate the problem of cooperative MARL as a decentralized partially observable Markov decision process (Dec-POMDP) \cite{oliehoek2016decpomdp}, defined as a tuple: 
\begin{equation}
\label{eqn:rb-dec-pomdp}
    \mathcal{G}=\langle \mathcal{N}, \mathcal{S}, \mathcal{O}, O, \mathcal{A}, \mathcal{P}, R, \gamma \rangle.
\end{equation}
Here $\mathcal{N}=\{1, ..., N\}$ is the set containing $N$ agents, $\mathcal{S}$ is the global state space, $\mathcal{O}=\times_{i\in\mathcal{N}} \mathcal{O}^i$ is the observation space, $O$ is the observation emission function, $\mathcal{A} = \times_{i \in \mathcal{N}}\mathcal{A}^i$ is the joint action space, $\mathcal{P}: \mathcal{S} \times \mathcal{A} \rightarrow \Delta(\mathcal{S})$ is the state transition probability, $R: \mathcal{S} \times \mathcal{A} \rightarrow \mathbb{R}$ is the shared reward function for cooperative agents, $\gamma \in [0, 1)$ is the discount factor.

At each timestep, agent $i$ observes $o^i_t = O(s_t, i)$ and add it to history $h^i_t = [o^i_0, a^i_0, ... ,o^i_{t}]$ to alleviate partial observability issue \cite{oliehoek2016decpomdp,rashid2018qmix}. Then, it takes action $a^i_t \in \mathcal A^i$ using policy $\pi^i(a^i_t|h^i_t)$. The joint actions $\mathbf a_t$ leads to the next state $s_{t+1}$ following state transition probability $P(s_{t+1}|s_{t}, \mathbf{a}_t)$ and shared global reward $r_t = R(s_t, \mathbf a_t)$. The objective for agents is to learn a joint policy $\pi(\mathbf a_t|\mathbf h_t) = \prod_{i \in \mathcal N} \pi^i(a_t^i|h_t^i)$ that maximize the value function $V_\pi(s) = \mathbb{E}_{s, \mathbf a}\left[\sum_{t=0}^{\infty} \gamma^t r_t | s_0 =s, \mathbf a_t \sim \pi(\cdot|\mathbf{h}_t) \right]$.

\paragraph{Robust Multi-Agent Reinforcement Learning.} Robust MARL aims to fortify against uncertainties in actions \cite{li2019M3DDPG}, states \cite{han2022staterobustmarl,he2023state}, rewards/environment \cite{kardecs2011robustsg,zhang2020rmaddpg,he2022envuncertainty}, and communications \cite{sun2022advcomm,xue2021advcomm}. Among these factors, action robustness have become a main focus due to the propensity for multiple agents to act unpredictably during deployment. Algorithms such as M3DDPG \cite{li2019M3DDPG} and ROMAX \cite{sun2022romax} treat each agent as an adversary that deviates towards jointly worst-case actions \cite{gleave2019iclr2020advpolicy,dinh2023nonoblivious}. However, in real world, since not all other agents are adversaries, such a policy can be overly pessimistic, or insufficiently robust. Later approaches attempt to directly train policies against these worst-case adversaries \cite{nisioti2021romq,phan2020robustmarl1,yuan2023evolution,li2023epr-mappo}. However, as these methods must explore numerous distinct adversarial scenarios, each scenario may left insufficiently examined. As a consequence, attackers can be less powerful comparing with worst-case adversary, and defenders trained with such weaker attackers can still be vulnerable to worst-case adversaries at test time.

\paragraph{Robustness without an Adversary.} While it is tempting to directly train MARL policy against adversaries via max-min optimization, such process can be overly pessimistic \cite{li2019M3DDPG}, insufficiently balanced across threat scenarios \cite{nisioti2021romq,li2023epr-mappo}, and computationally demanding \cite{derman2021twiceregularized}. A parallel line of research in RL aims to achieve robustness without relying on adversaries. A2PD \cite{qu2021a2pd} shows a certain modification of policy distillation can be inherently robust against state adversaries. Through the use of convex conjugate, \cite{eysenbach2021maxentrl} have shown that max-entropy RL can be provably robust against uncertainty in reward and environment transitions. \cite{derman2021twiceregularized} further extended regularization to uncertainties in reward and transition dynamics under rectangular and ball constraints. The work most similar to ours is ERNIE \cite{bukharin2024ernie}, which minimize the Lipshitz constant of value function under worst-case perturbations in MARL. However, the method considers all agents as potential adversaries, thus inherits the drawback of M3DDPG, learning policy that can either be pessimistic or insufficiently robust.

\section{Method}


Unlike current robust MARL approaches that prepares against every conceivable threat, human learns in routine scenarios, but can reliably reflect to all types of threats encountered. Drawing inspiration from human adaptability to unseen threats, we first formalize robust MARL as an action adversarial Dec-POMDP, aiming to maximize both cooperative and robust performance under all threat scenarios. Our approach frames this as an inference problem \cite{levine2018probinf}, where policies are learned in a Dec-POMDP without attack, and adapts to diverse worst-case scenarios using off-policy evaluation. We find that minimizing mutual information between histories and actions maximizes a lower bound for robustness. Beyond theoretical derivations, our method not only acts as an information bottleneck to reduce spurious correlations but also facilitates the learning of robust action priors, which better maintains effective tactics even under attack.


\subsection{Problem Formulation}

\textbf{Action Adversarial Dec-POMDP.} In this paper, we consider action uncertainty as an unknown portion of agents taking unexpected actions. This can stem from robots losing control due to software/hardware error, or are compromised by an adversary \cite{gleave2019iclr2020advpolicy,nisioti2021romq,zhang2020rmaddpg,li2023epr-mappo,yuan2023evolution}. Given a Dec-POMDP with action uncertainties, we define action uncertainties in MARL as an action adversarial Dec-POMDP (A2Dec-POMDP), which is written as:
\begin{equation}
\label{eqn:a2dec-pomdp}
    \hat{\mathcal{G}}=\langle \mathcal{N}, \Phi, \mathcal{S}, \mathcal{O}, O, \mathcal{A}, \mathcal{P}, R, \gamma \rangle.
\end{equation}
Here $\Phi = \{0, 1\}^N$ is a set containing \emph{partitions} of agents into defenders and adversaries, with $\phi \in \Phi$ indicates a specific partition. For each agent $i$, $\phi^i = 1$ means the original policy of $\pi^i(\cdot|h_t^i)$ is replaced by a worst-case adversarial policy $\pi^i_\alpha(\cdot|h_t^i, \phi)$, while $\phi^i = 0$ means the original policy is executed without change. In this way, Dec-POMDP is a special case of A2Dec-POMDP with $\phi = \mathbf 0_N$.

\textbf{Perturbed policy.} The perturbed joint policy is defined as $\hat{\pi}(\hat{\mathbf a}_t|\mathbf h_t, \phi) = \prod_{i \in \mathcal N} [\pi^i_\alpha(\cdot|h_t^i, \phi) \cdot \phi + \pi^i(\cdot|h_t^i) \cdot (1 - \phi)]$, with perturbed joint actions $\hat{\mathbf a}_t$ used for environment transition $\mathcal P(s_{t+1}|s_t, \hat{\mathbf a}_t)$ and reward $r_t = R(s_t, \hat{\mathbf a}_t)$. For each $\phi \in \Phi$, the value function is $V_{\pi, \pi_\alpha}(s, \phi) = V_{\hat{\pi}}(s, \phi) = \mathbb{E}_{s, \hat{\mathbf a}}\left[\sum_{t=0}^{\infty} \gamma^t r_t | s_0 =s, \hat{\mathbf a}_t \sim \hat{\pi}(\cdot|\mathbf h_t, \phi) \right]$.

\textbf{Attacker's objective.} We assume the attack happens at test time, with parameters in defender's policy $\pi$ fixed during deployment. For a partition $\phi$ that indicates defenders and adversaries, the objective of a worst-case, zero-sum adversary aims to learn a joint adversarial policy $\pi_{\alpha}(\cdot|\mathbf{h}_t, \phi) = \prod_{i \in \{\phi^i = 1\}} \pi^i_{\alpha}(\cdot|h_t^i, \phi)$ that minimize cumulative reward \cite{gleave2019iclr2020advpolicy,dinh2023nonoblivious}:
\begin{equation}
\begin{split}
\label{eqn:perturb}
    \pi_{\alpha}^{*} \in \argmin\limits_{\pi_{\alpha}} V_{\pi, \pi_\alpha}(s, \phi).
\end{split}
\end{equation}
Following \cite{gleave2019iclr2020advpolicy}, an optimal worst-case adversarial policy $\pi_{\alpha}^{*}$ always exists for all possible partitions $\phi \in \Phi$ and fixed $\pi$. Since the defender's policy $\pi$ is held fixed during attack, we can view it as a part of environment transition, reducing the problem to a POMDP for one adversary or a Dec-POMDP for multiple adversaries. The existence of an optimal $\pi_{\alpha}^{*}$ is then a corollary of the existence of an optimal policy in POMDP \cite{aastrom1965pomdp} and Dec-POMDP \cite{oliehoek2016decpomdp}.

\textbf{Defender's objective.} The objective of defenders is to learn a policy that maximize both normal performance and robust performance under attack, without knowing who is the adversary:
\begin{equation}
\label{eqn:minmax}
    \pi^{*} \in \argmax\limits_{\pi} \bigg[ V_{\pi}(s) + \mathbb{E}_{\phi \in \Phi^\alpha} \left[\min\limits_{\pi_{\alpha}} V_{\pi, \pi_\alpha}(s, \phi) \right] \bigg].
\end{equation}
Here we use $\Phi^\alpha = \Phi \backslash \mathbf{0}_N$ to denote a set of partitions that contains at least one adversary. While existing max-min approaches requires explicitly training $\pi$ with all $\phi \in \Phi$, our method trains $\pi$ with partition $\phi = \mathbf 0_N$ only, but still capable of solving the max-min objective in Eqn. \ref{eqn:minmax}. This is done by deriving a lower bound for objective $\min\limits_{\pi_{\alpha}} \mathbb{E}_{\phi \in \Phi^\alpha} \left[V_{\pi, \pi_\alpha}(s, \phi) \right]$ as a regularization term.

\subsection{MIR3 is Provably Robust}

We adopt a control-as-inference approach \cite{levine2018probinf} to infer the defender's policy $\pi$. We first derive objectives for purely cooperative scenarios, then show objectives under attack by importance sampling. Let $\tau^0 = [(s_0, \mathbf{a}_0), (s_1, \mathbf{a}_1), ... (s_t, \mathbf{a}_t)]$ denote the \emph{optimal} trajectory of purely cooperative scenario generated on $t$ consecutive stages, with superscript in $\tau^0$ denotes $\phi = \mathbf 0_N$. Following \cite{levine2018probinf}, the probability of $\tau$ being generated is:
\begin{equation}
\begin{split}
\label{eqn:perturb}
    p(\tau^0) = \left[ p(s_0) \prod_{t=0}^T \mathcal P(s_{t+1}|s_t, \mathbf{a}_t)\right] \exp \left(\sum_{t=1}^T r_t \right),
\end{split}
\end{equation}
with $\exp(\sum_{t=1}^T r_t)$ encourage trajectories with higher rewards to have exponentially higher probability \cite{levine2018probinf}. The goal is to find the best approximation of joint policies $\pi(\mathbf a_t|\mathbf h_t) = \prod_{i \in \mathcal N} \pi^i(a_t^i|h_t^i)$, such that its induced trajectories $\hat{p}(\tau^0)$ match the optimal probability of $p(\tau^0)$:
\begin{equation}
\begin{split}
\label{eqn:perturb}
    \hat{p}(\tau^0) =  p(s_0) \Bigg[ \prod_{t=0}^T \mathcal P(s_{t+1}|s_t, \mathbf{a}_t) \pi(\mathbf a_t|\mathbf h_t)\Bigg].
\end{split}
\end{equation}
Assume the dynamics is fixed, such that agents cannot influence the environment transition probability \cite{wen2019pr2}, the objective for purely cooperative scenario is derived as maximizing the negative of KL divergence between sampled trajectory $\hat{p}(\tau^0)$ and optimal trajectory $p(\tau^0)$:
\begin{equation}
\begin{split}
\label{eqn:perturb}
J^{0}(\pi) &= -D_{KL}(\hat{p}(\tau^0)||p(\tau^0)) = \textstyle{\sum_{t=1}^T} \mathbb E_{\tau^0 \sim \hat{p}(\tau^0)} [r_t + \mathcal H(\pi(\mathbf a_t|\mathbf h_t))],
\end{split}
\end{equation}
where $\mathcal H(\pi(\mathbf a_t|\mathbf h_t))$ is the entropy of joint policy.

As for scenarios with attack, for partition $\phi \in \Phi^\alpha$, let $\tau^\phi = [(s_1, \hat{\mathbf{a}}_1), (s_2, \hat{\mathbf{a}}_2), ... (s_t, \hat{\mathbf{a}}_t)]$ denote the trajectories under attack. To evaluate the performance of $\pi$ with partition $\phi$, we can leverage importance sampling to derive an unbiased estimator $J^{\phi}(\pi)$ using $\tau^0$, with $\rho_t = \frac{\hat{\pi}(\mathbf a_t|\mathbf h_t, \phi)}{\pi(\mathbf a_t|\mathbf h_t)}$ the per-step importance sampling ratio:
\begin{equation}
\begin{split}
\label{eqn:perturb}
J^{\phi}(\pi) & =\textstyle{\sum_{t=0}^T} \mathbb E_{\tau^0 \sim \hat{p}(\tau^0)} \left[ \rho_t  \cdot\left(r_t + \mathcal H(\pi(\mathbf a_t|\mathbf h_t)) \right)\right] = \textstyle{\sum_{t=0}^T} \mathbb E_{\tau^\phi \sim \hat{p}(\tau^\phi)} \left[r_t + \mathcal H(\pi(\mathbf a_t|\mathbf h_t))\right].
\end{split}
\end{equation}
Following Eqn. \ref{eqn:minmax}, we can now derive the overall objective $J(\pi)$ for inference:
\begin{equation}
\begin{split}
\label{eqn:perturb}
J(\pi) & = J^0(\pi) + \mathbb{E}_{\phi \in \Phi^\alpha} \left[ \min\limits_{\pi_{\alpha}} J^\phi(\pi) \right] \\
&= \textstyle{\sum_{t=0}^T} \mathbb E_{\tau^0 \sim \hat{p}(\tau^0)} [r_t + \mathcal H(\pi(\mathbf a_t|\mathbf h_t))]  + \mathbb{E}_{\phi \in \Phi^\alpha} \big[ \min\limits_{\pi_{\alpha}} \textstyle{\sum_{t=0}^T}  \mathbb E_{\tau^\phi \sim \hat{p}(\tau^\phi)} \left[r_t + \mathcal H(\pi(\mathbf a_t|\mathbf h_t))\right]\big].
\end{split}
\end{equation}

Thus, our objective aims to maximize cumulative reward both cooperative scenarios and across all possible defender-adversary partitions (\ie, threat scenarios), denoted by $\phi \in \Phi^\alpha$, through off-policy evaluation. We now present our main result.

\begin{proposition}
\label{jointdev}
\emph{$J(\pi) \geq \sum_{t=1}^T \mathbb E_{\tau^0 \sim \hat{p}(\tau^0)}[r_t - \lambda I(\mathbf{h}_t; \mathbf{a}_t)]$, where $\lambda$ is a hyperparameter\footnote{In principle, we do not need $\lambda$ since it can be absorbed into reward function. Here we make it explicit to represent the tradeoff between reward and mutual information, which is standard in literature \cite{haarnoja2017sql}.}.}
\end{proposition}
\emph{proof sketch:} The proof proceeds in three steps. First, we transform the benign policy into an adversarial one using probabilistic inference. Second, we derive a lower bound for trajectories that include adversaries. Finally, we translate this lower bound into the expression $-I(\mathbf h_t; \mathbf{a}_t)$. \emph{A complete proof can be found in Appendix. \ref{appendix_proof}.} \qed

The relation between minimizing the objective $I(\mathbf{h}_t; \mathbf{a}_t)$ and enhancing robustness is intuitive. When some agents fail due to uncertainties, their erroneous actions will alter the global state, affecting future observations and ultimately the histories of other benign agents. Compared to the intuitive approach of minimizing the mutual information between agents' actions, our objective also accounts for environmental transitions under the control-as-inference framework.

Finally, all we need is to add the mutual information between histories and joint actions $- \lambda I(\mathbf h_t; \mathbf{a}_t)$ as a robust regularization term to reward $r_t$. Since our MIR3 is only an additional reward, it can be optimized by \emph{any} cooperative MARL algorithms. Technically, the exact value of $I(\mathbf h_t; \mathbf{a}_t)$ is intractable to calculate, so we estimate its upper bound as a lower bound for $-I(\mathbf h_t; \mathbf{a}_t)$. We use CLUB \cite{cheng2020club, li2022pmicinfluenceicml}, an off-the-shelf mutual information upper bound estimator, to estimate this information. \emph{A pseudo code of our MIR3 is given in Appendix. \ref{appendix_algo}.}



\subsection{Insights and Discussions}

\begin{figure}[t]
    \begin{minipage}[t]{0.49\textwidth}
        \centering
        \includegraphics[width=\linewidth]{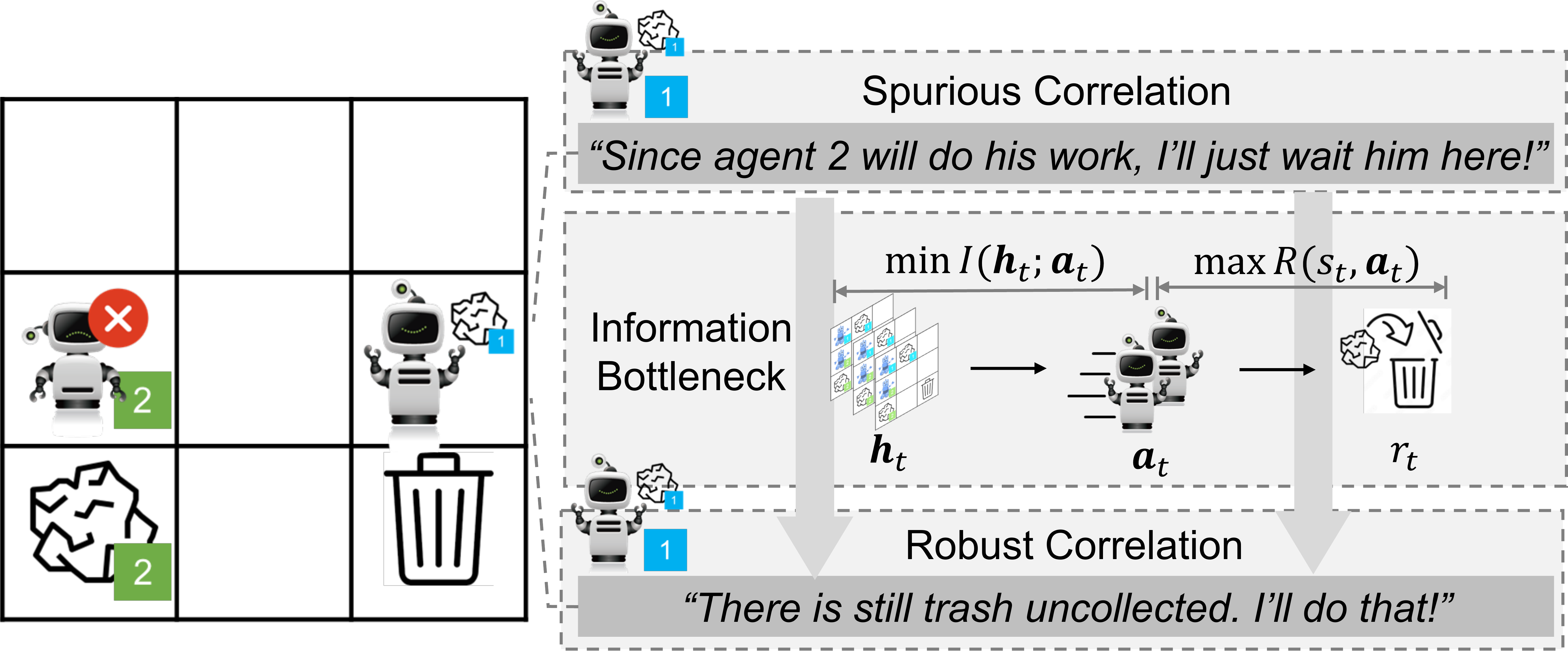}
        \captionof{figure}{MIR3 as information bottleneck, eliminating spurious correlations in histories and mitigating overreactions to agents with action uncertainties, forming robust agent-wise interactions.}
        \label{informationbottleneck}
    \end{minipage}
    \hfill
    \begin{minipage}[t]{0.49\textwidth}
        \centering
        \includegraphics[width=0.8\linewidth]{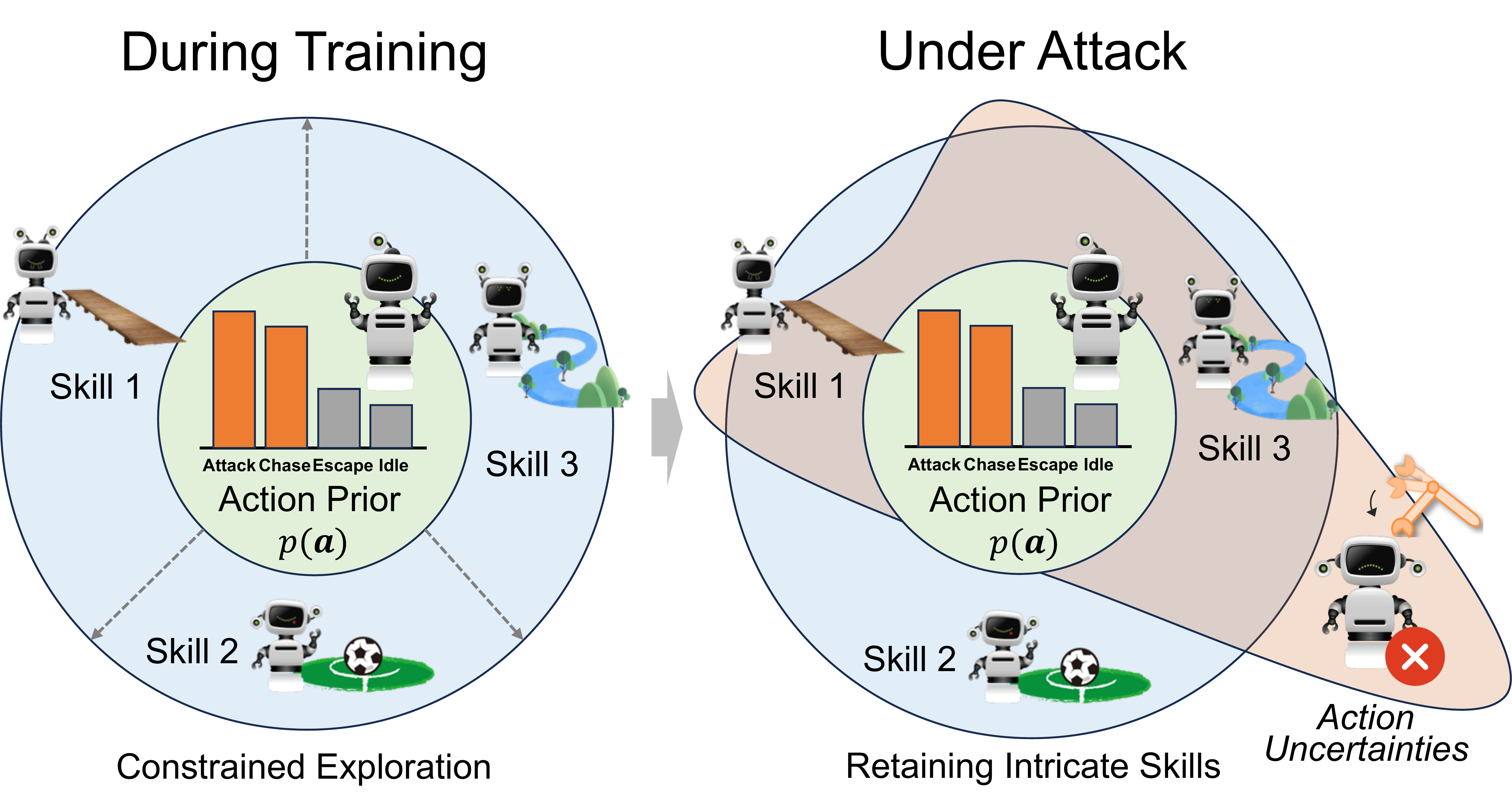}
        \captionof{figure}{MIR3 as robust action prior. The objective bias policy to effective actions in the environment, and fosters exploration around this action prior to handle task variations and uncertainties.}
        \label{prior}
    \end{minipage}
    \vspace{-0.2in}
\end{figure}

Beyond theoretical basis, our MIR3 can be seen as an information bottleneck that reduce unnecessary correlations between agents, or as learning a robust action prior that favors effective actions in the environment. These discussions provide explanations for the success of our approach.

\textbf{MIR3 as Information bottleneck.} Our mutual information minimization objective can be seen as an information bottleneck, which encourage policies to eliminate spurious correlations among agents. This concept, initially introduced by Tishby \etal, seeks to identify a compressed representation that retains the maximum relevant information with the label \cite{tishby2000ib,tishby2015ib,shwartz2017ib,saxe2019ib}. In MARL, as depicted in Fig. \ref{informationbottleneck}, our objective $\max_{\pi} \mathbb E_{\tau^0 \sim p(\tau^0)} \left[r_t - \lambda I(\mathbf h_t; \mathbf{a}_t) \right]$ functions as an information bottleneck, considering history as input, actions as an intermediate representation, and reward as the final label. The aim is to find a set of actions employing \emph{minimal sufficient information} from the current history, which is maximally relevant for solving the task and getting higher reward.

The objective is crucial for eliminating spurious correlation between agents, which helps handling action uncertainties in MARL. For example, robot swarms trained in simulation environment assumes each agent to be optimally cooperative to enable best performance. As shown in Fig. \ref{informationbottleneck}, this objective can form a spurious correlation that encourage robots to \emph{overly} rely on others. In reality, individual robots can malfunction due to software/hardware errors, execute suboptimal actions, or send erroneous signals, which is reflected in histories. As such, information bottleneck encourage agents not to overly rely on current history, and form a loose cooperation with others only in case of need. Therefore, even if some agents falter, our objective enables the remaining agents to fulfill their tasks independently without overreacting or being swayed by failed agents.

\textbf{MIR3 as robust action prior.} Minimizing mutual information can also be seen as a robust action prior, which favors useful actions conditioned on current task and maintains intricate tactics under action uncertainties via exploration. In information theory, $- I(\mathbf h; \mathbf a) = \mathbb E_{\mathbf h \sim p(\mathbf h)}[- D_{KL}(\pi(\mathbf a|\mathbf h) || p(\mathbf a))]$, thus ensuring the exploration of the policy does not diverge significantly from marginal distribution $p(\mathbf a)$. This concept aligns with the concept of \emph{action prior} in literature \cite{grau2018actionprior,pertsch2021actionprior}. Similarly, the widely used max entropy RL objective \cite{eysenbach2021maxentrl}, $\mathcal H(\mathbf a|\mathbf h)$, can be seen as using a uniform action prior.


As shown in Fig. \ref{prior}, the benefit of constraining a policy to $p(\mathbf a)$ on robustness can be interpreted from two aspects. First, $p(\mathbf a)$ can be viewed as a set of task-relevant actions consistently favored by the environment, independent of current histories. For example, in StarCraft II, actions directed at moving towards and attacking enemies are usually preferred for victory. More intricate tactics, such as kiting or focused fire, are optional and depend on current histories \cite{samvelyan2019smac}. Thus, if certain actions are broadly effective within the environment, the policy is prone to succeed in accomplishing the task by leaning on these actions, even when confronted with action uncertainties. Secondly, keeping the policy near $p(\mathbf a)$ fosters exploration in its vicinity. Therefore, even if some agents deviate from the optimal policy, the enhanced exploration around $p(\mathbf a)$ encourages the policy to identify diverse methods for handling the task, preserving some intricate tactics for the task to succeed.

\section{Experiments}

\subsection{Experiment settings}

\textbf{Environments.} We evaluated our result on six tasks in StarCraft Multi-agent Challenge (SMAC) \cite{samvelyan2019smac} and a continuous continuous robot swarm control task with 10 agents performing \emph{rendezvous}, which agents are randomly placed in the arena and learns to gather together. In all tasks, agents are required to complete the task with worst-case adversaries during testing, which differs from standard cooperative MARL setting. For SMAC, we find having adversary controlling one agent makes the environment unsolvable. We address this by allowing algorithms control over additional agents to ensure fair evaluation. For rendezvous, we additionally deploy the trained algorithm on real robots, and test its performance in a real world arena.

\textbf{Compared methods.} We evaluate the performance of MIR3 on MADDPG \cite{lowe2017maddpg} and QMIX \cite{rashid2018qmix} backbones. The compared methods includes M3DDPG \cite{li2019M3DDPG}, ROMAX \cite{sun2022romax} and ERNIE \cite{bukharin2024ernie}, which considers all other agents as adversaries; ROM-Q \cite{nisioti2021romq} which considers the performance of agents in each threat scenarios. Note that the design of M3DDPG \cite{li2019M3DDPG} and ROMAX \cite{sun2022romax} relies on the central critic of MADDPG, so we do not evaluate it on QMIX backbone. All methods are compared based on the same network architecture, hyperparameters and tricks. \emph{We leave hyperparameters and implementation details in Appendix. \ref{appendix_details}. See code and demo videos in Supplementary Materials.}

\textbf{Evaluation protocol.} For environment with $N$ agents, all methods to be attacked was trained using five random seeds. During attack, we fix the parameters in defender's policy, and train a worst-case adversary against current policy \cite{gleave2019iclr2020advpolicy, dinh2023nonoblivious}. For scenarios with one agent as adversary, we average the attack result of each N agents using the same five seed, reporting results averaged on $5*N$ seeds. For scenarios with more than one adversary, we randomly sample 5 attack scenarios and report the result using the same five seed. All results are reported with 95\% confidence interval.

\begin{figure*}[t]
\centering
\includegraphics[scale=0.22]{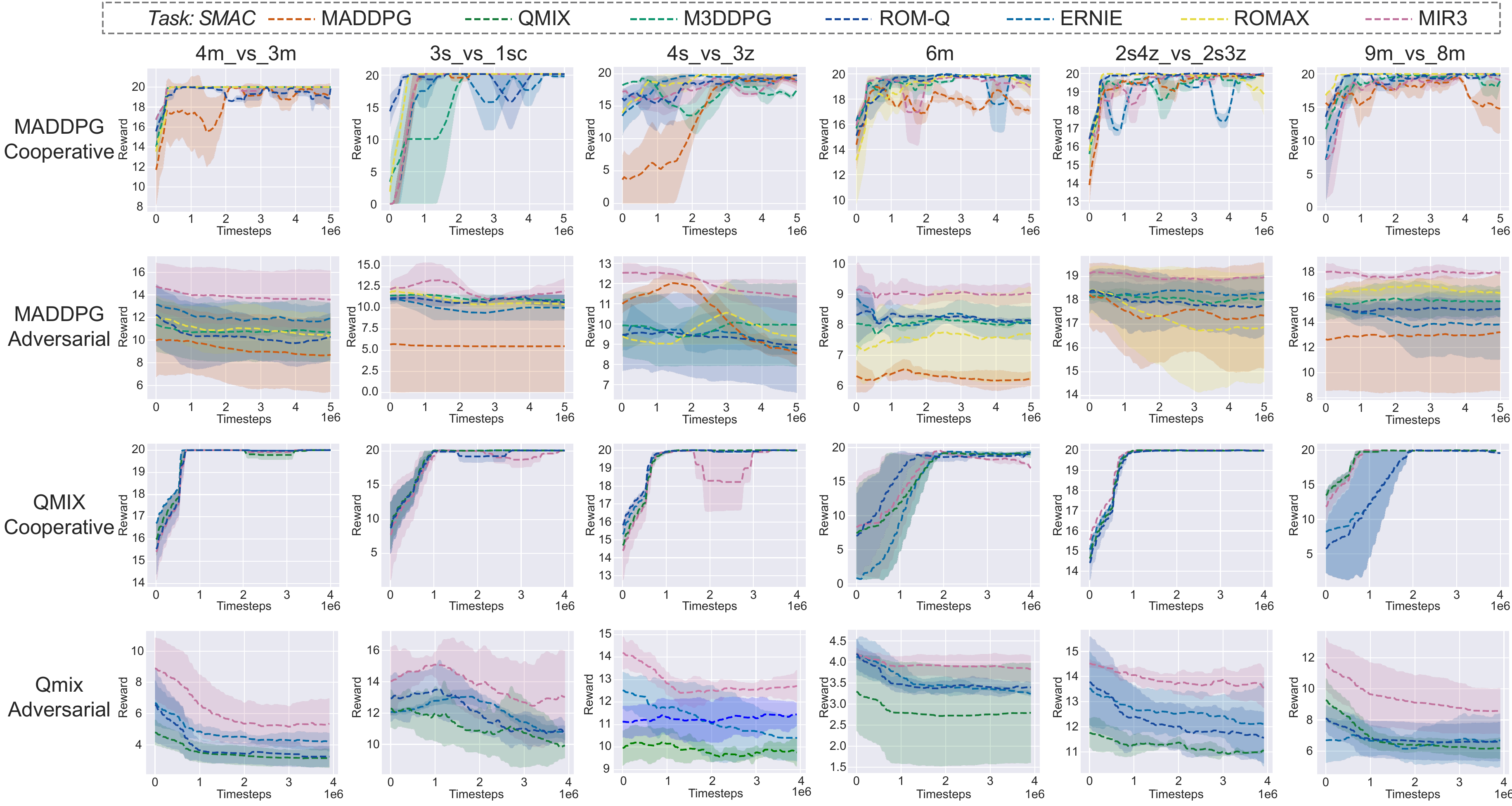}
\caption{Cooperative and robust performance on six SMAC tasks, evaluated on MADDPG and QMIX backbones. While never seen adversaries, our MIR3 approach outperforms baselines that explicitly consider threat scenarios. Results reported on 5 seeds for cooperative and $5 \times N$ seeds for adversary scenarios with 95\% confidence interval.}
\label{mainexp}
\vspace{-0.1in}
\end{figure*}

\begin{figure*}[t]
\centering
\includegraphics[scale=0.18]{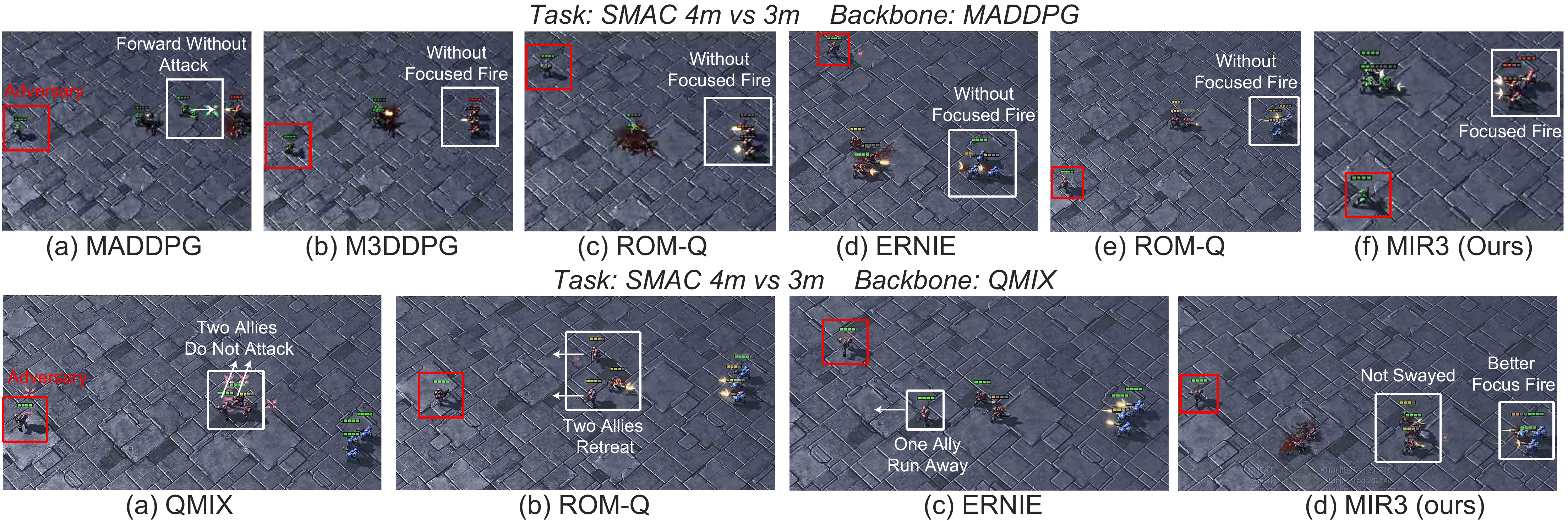}
\caption{Agent behaviors under attack in task \emph{4m vs 3m}, adversary denoted by red square. Under MADDPG backbone, baselines are either swayed by adversaries or lack cooperation on focused fire. Under QMIX backbone, baselines are frequently swayed without attack. In contrast, our MIR3 is now swayed by adversary and preserves cooperation on focused fire.}
\label{visual}
\vspace{-0.1in}
\end{figure*}

\subsection{Simulation Results}

We first present our results on six SMAC tasks. Experiments show our MIR3 significantly surpasses baselines in robustness and training efficiency, while maintaining cooperative performance. The result of multi-agent rendezvous will be discussed in next subsection.

\textbf{MIR3 is more robust.} We evaluate the defense capability of MIR3 against worst-case attacks, with one agent as an adversary. Experiments involving more adversaries will be discussed later. As shown in Fig. \ref{mainexp}, although MIR3 does not encounter adversaries during training, it demonstrates superior defense capabilities across six tasks and two backbones, consistently outperforming even the best-performing baselines that directly consider adversaries.

The improved performance of MIR3 over baselines can be explained as follows. Compared to M3DDPG, ERNIE, and ROMAX, which assume all other agents as potential adversaries, MIR3 avoids learning overly pessimistic or less effective policies; Compared to ROM-Q which prepares for each threat scenario, our approach shows that when adversaries and defenders cannot adequately explore or respond to the myriad threat scenarios, the adversaries remain weak during training, leading to less effective defenders. Overall, our results in Appendix. \ref{appendix_romq} demonstrate that while all baselines are effective against uncertainties during training, their defenses can be easily compromised by worst-case adversaries at test time. In contrast, MIR3, without exploring any threat scenarios, implicitly maximizes the lower bound performance under any threat scenario.


\textbf{MIR3 does not harm cooperative performance.} We further show our MIR3 maintains cooperative performance while enhancing robustness. This is achieved by minimizing mutual information as an information bottleneck, which has been reported to enhance task performance in computer vision tasks \cite{alemi2016ibrobust}. Additionally, this is supported by the objective in Eqn. \ref{eqn:minmax}, where defenders maximize both cooperative and robust performance.

\textbf{MIR3 learns robust behaviors.} Next, we show that MIR3 learns distinct robust behaviors. As illustrated in Fig. \ref{visual}, under the MADDPG backbone, vanilla MADDPG without defense can be easily swayed by adversaries, causing agents to move forward without attacking and getting killed by enemies. Other robust baselines are rarely swayed but fail to retain cooperative behaviors (\eg, no focused fire on enemies), eventually losing the game. In contrast, by reducing mutual information, MIR3 ensures that agents are not only unswayed but also maintain focused fire behavior under attack.

Under the QMIX backbone, benign agents in all baselines are frequently swayed by the adversary, moving randomly without attacking. In contrast, MIR3 agents are less swayed by the adversary and perform better focused fire than the baselines. Notably, QMIX agents generally perform worse than MADDPG agents. We hypothesize that this discrepancy arises because QMIX assumes all agents contribute positively to the team, an assumption that does not hold in the presence of adversaries. \emph{See additional analysis on task \emph{9m vs 8m} in Appendix. \ref{appendix_smac_video} and videos in Supplementary Materials.}



\begin{figure}[t]
\centering
\includegraphics[scale=0.36]{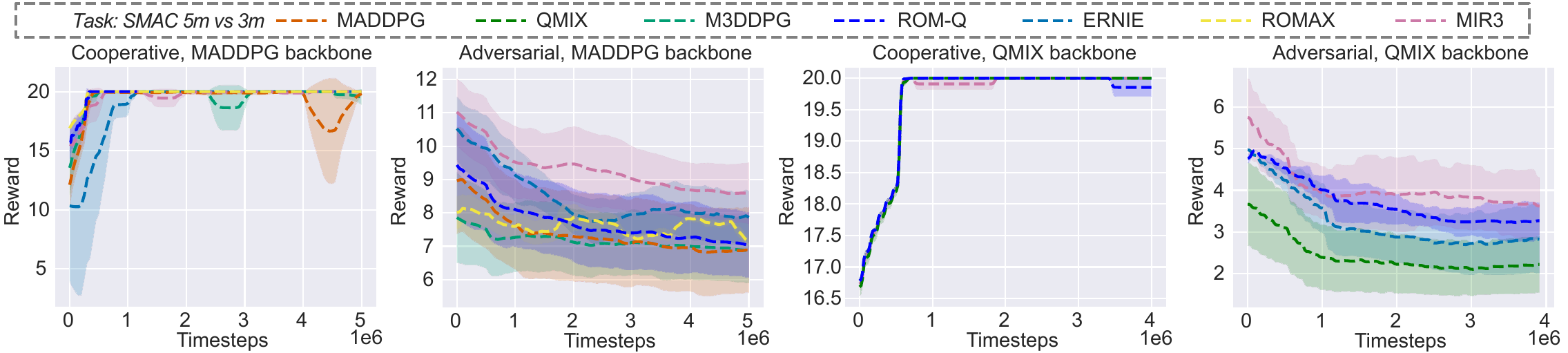}
\caption{Defense with two adversaries, evaluated in SMAC \emph{5m vs 3m}. Our MIR3 is consistently more robust comparing with baselines. \emph{See results of another five SMAC tasks in Appendix. \ref{appendix_twoadv}.}}
\label{twoadv}
\vspace{-0.25in}
\end{figure}

\textbf{MIR3 is robust with many adversaries.} In extreme situations, there could be more than one adversaries. We examined this by adding an extra adversarial agent in map \emph{4m vs 3m} in SMAC, creating a map \emph{5m vs 3m} with two adversaries. As illustrated in Fig. \ref{twoadv}, in this challenging scenario, our MIR3 consistently exhibits stronger defense capability than all baselines, in both MADDPG and QMIX backbones. This demonstrates the potential of our MIR3 to be applied in more complex scenarios with many adversaries. \emph{See results of another five SMAC tasks in Appendix. \ref{appendix_twoadv}.}

\begin{minipage}{0.48\textwidth}
    \centering
    \tiny
    \setlength{\tabcolsep}{1.5mm}
    \begin{tabular}{cccc}
    \hline
                & SMAC-MADDPG      & SMAC-QMIX    & Rendezvous  \\ \hline
    MADDPG      & $0.28\pm 0.11$ & -   & $0.61\pm 0.04$   \\ \hline
    QMIX      & - & $0.69\pm 0.17$   & -   \\ \hline \hline
    M3DDPG      & $0.41\pm 0.12$  & - & $2.16\pm 0.29$   \\ \hline
    ROM-Q       & $0.42\pm 0.15$ & $1.01\pm 0.08$   & $2.43\pm 0.29$ \\ \hline
    ROMAX      & $0.48\pm 0.14$ & -   & $2.82\pm 0.40$ \\ \hline
    ERNIE      & $0.40\pm 0.14$ & $0.98\pm 0.08$   & $1.57\pm 0.12$ \\ \hline
    MIR3 (Ours) & $\mathbf{0.31}\pm 0.16$  & $\mathbf{0.81}\pm 0.09$   & $\mathbf{0.63}\pm 0.04$  \\ \hline
    \end{tabular}
\captionof{table}{Per epoch training time of our MIR3 and baselines, reported in second. MIR3 only adds little training time to MADDPG and QMIX backbones, while much faster than methods that considers threat scenarios explicitly.}
\label{time}
\vspace{0.1in}
\end{minipage}%
\hfill 
\begin{minipage}{0.48\textwidth}
\centering
\includegraphics[width=0.9\linewidth]{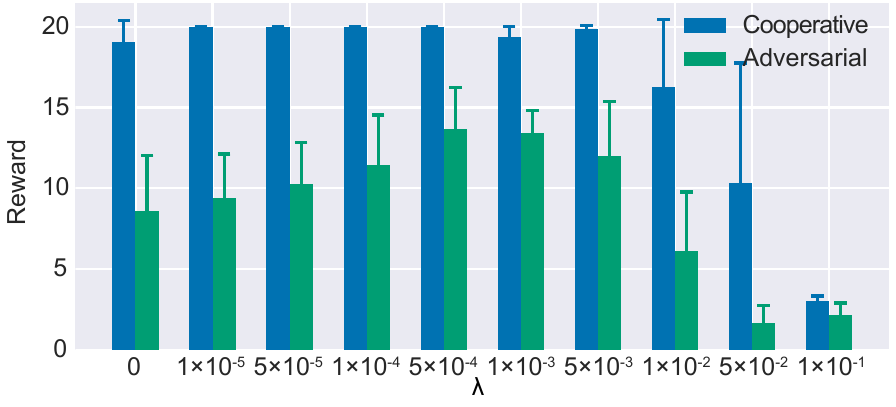}
\captionof{figure}{Ablations on hyperparameter $\lambda$ that suppress mutual information, showing a tradeoff between policy effectiveness and limiting information flow. Evaluated on SMAC \emph{4m vs 3m}.}
\label{ablation}
\vspace{0.1in}
\end{minipage}


\textbf{MIR3 requires less training time.} We also demonstrate that our MIR3 method is computationally more efficient than baselines that explicitly consider threat scenarios. Following \cite{qu2021a2pd}, we report the average training time per epoch over 50 episodes. All statistics are obtained based on one Intel Xeon Gold 5220 CPU and one NVIDIA RTX 2080 Ti GPU, using task \emph{4m vs 3m} for SMAC and \emph{10 agents} for rendezvous. As shown in Table. \ref{time}, our MIR3 only requires slightly more training time than backbones without considering robustness (+10.71\% in SMAC-MADDPG, +17.39\% in SMAC-QMIX, +3.28\% in rendezvous), showing our defense can be added at low cost. In contrast, considering threat scenarios involves the costly approach of approximating an adversarial policy, resulting in significantly higher training times compared to our MIR3 approach (+29.03\% in SMAC-MADDPG, +20.99\% in SMAC-QMIX, +149.21\% in rendezvous).

\textbf{Ablations on hyperparameters.} Finally, we study the effect of hyperparameter $\lambda$ in penalizing mutual information between histories and actions, which can be seen as an information bottleneck. We set $\lambda$ in $\{0, 10^{-5}, ... , 10^{-1}\}$ and evaluate the result on task \emph{4m vs 3m} in SMAC with MADDPG backbone. The results are illustrated in Fig. \ref{ablation}, note that with $\lambda = 0$, our MIR3 reduces to MADDPG.


For relatively small $\lambda$ (\ie, $\lambda \leq 5\times10^{-4}$), the policy is steered to focus less, but more relevant information in current history. This efficiently suppresses unnecessary agent-wise interactions, leading to more robust policies and even slightly enhancing cooperative performance, which is also evident in computer vision tasks using information bottleneck as regularizer \cite{alemi2016ibrobust}. Conversely, when $\lambda> 5\times10^{-4}$, the policy is restricted from utilizing any information from the current history, resulting in a collapse of both cooperative and robust performance. As a consequence, we select $\lambda = 5\times10^{-4}$ for an optimal tradeoff between limiting information flow and maintaining policy effectiveness.

\subsection{Real World Experiments}

\begin{figure}[t]
    \centering
    \begin{subfigure}[t]{0.24\textwidth}
         \centering
         \includegraphics[width=\textwidth]{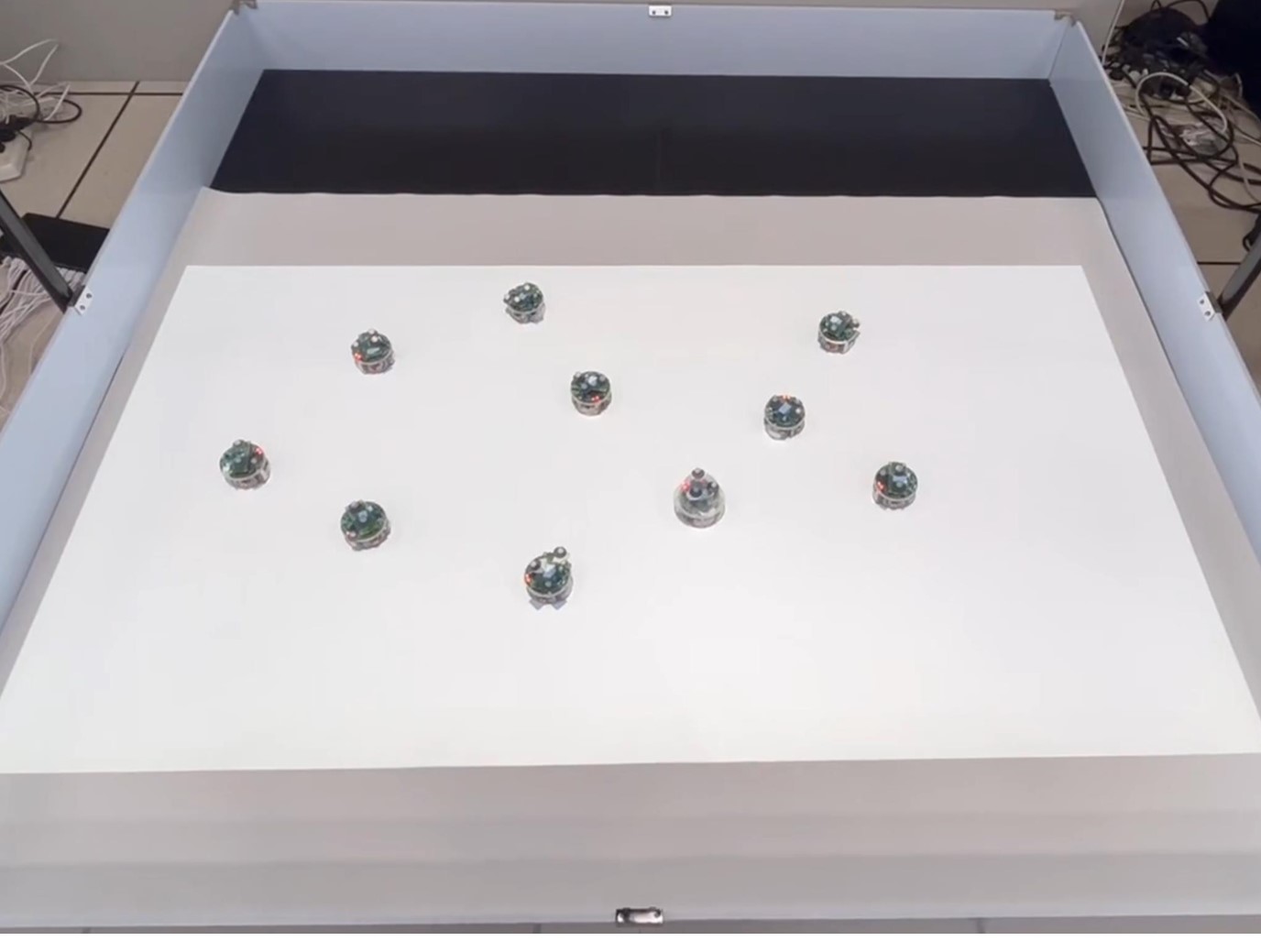}
         \caption{Arena}
         \vspace{-0.05in}
         \label{arena}
     \end{subfigure}
     \hfill
     \begin{subfigure}[t]{0.47\textwidth}
         \centering
         \includegraphics[width=\textwidth]{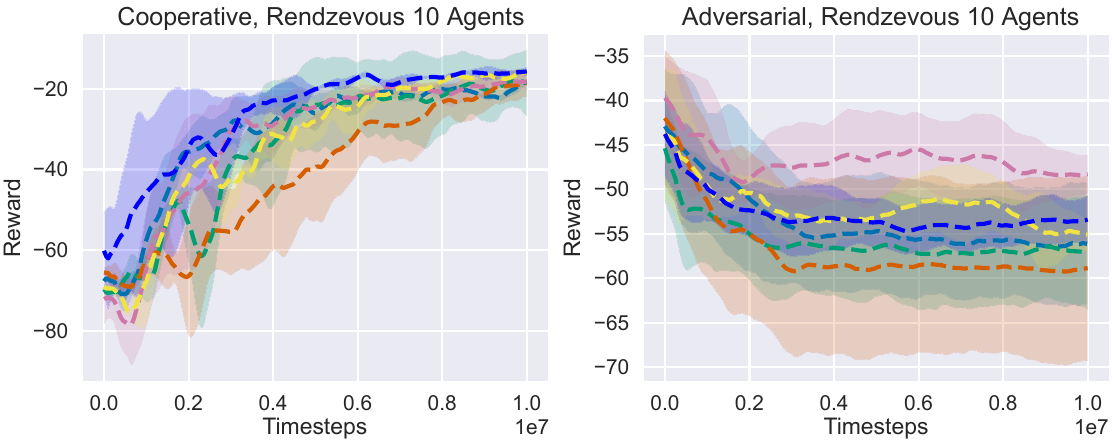}
         \caption{Simulation Results}
         \label{realworld_sim}
         \vspace{-0.05in}
     \end{subfigure}
     \hfill
    \begin{subfigure}[t]{0.24\textwidth}
         \centering
         \includegraphics[width=\textwidth]{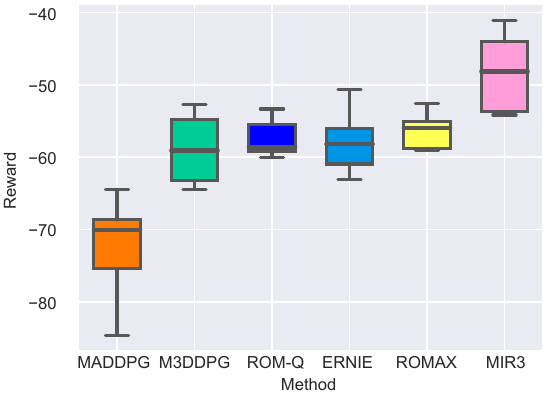}
         \caption{Real World Results}
         \label{phys-result}
         \vspace{-0.05in}
     \end{subfigure}
    \caption{Illustration of our real-world rendezvous environment and results against action uncertainties in both simulation and real world deployment.}
    \label{phys_exp}
    \vspace{-0.1in}
\end{figure}

In this section, we evaluate the robustness of our MIReg under action uncertainties in real world robot swarm control, with unknown environment conditions, inaccurate control dynamics and noisy sensory input. As shown in Fig. \ref{arena}, our experiments are conducted in an $2m \times 2m$ indoor arena with 10 e-puck2 robots \cite{mondada2009epuck}. In alignment with the widely accepted sim2real paradigm in the reinforcement learning community \cite{hofer2021sim2real}, we directly transfer the policies for both defenders and adversaries trained in simulation environments, to our robots in real world.

\begin{figure*}[t]
\centering
\includegraphics[scale=0.23]{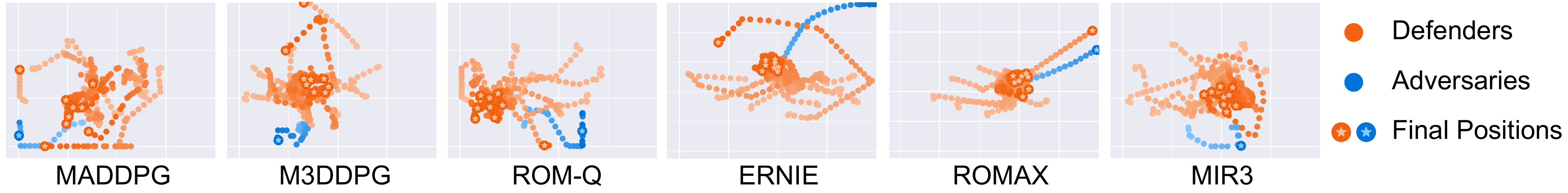}
\caption{Real-world trajectories of robot swarm performing \emph{rendezvous}, with defenders in orange and adversaries in blue. We use a star to denote the final position. Our MIR3 agents reliably perform the task of getting together without being swayed by adversary. \emph{See videos in Supplementary Materials.}}
\label{phys-traj}
\vspace{-0.2in}
\end{figure*}

The results are organized as follows. First, in simulation, our MIR3 consistently outperforms baselines in robustness, without sacrificing cooperative performance. It is interesting to note that in our simulation, while only trained on \emph{rendezvous} task, our MIR3 agents shows an emergent pursuit-evade behavior when facing adversary running away. \emph{See depictions of this behavior in Appendix. \ref{appendix_video} and videos in Supplementary Materials.}

Second, when deployed in real world, our MIR3 consistently shows greater resilience. As illustrated in Fig. \ref{phys-result}, our MIR3 achieves +14.29\% average reward improvement compared to the best performing baseline. Moreover, as shown in Fig. \ref{phys-traj}, a detailed examination of the trajectories reveals that MIR3 successfully learns robust behaviors. In contrast to simulation, MADDPG completely failed to handle real-world uncertainties, leading to multiple agents malfunctioning and failing to gather, underscoring the necessity of evaluating robustness in real world. M3DDPG, ROM-Q, ERNIE and ROMAX perform substantially better than MADDPG, although one or several agents are still misled by the adversary. Conversely, our MIR3 can group together without deviation, and maintain consistent behavior throughout the evaluation. \emph{See videos in Supplementary Materials.}

\section{Conclusions}

In this paper, we introduce MIR3, a novel regularization-based approach for robust MARL. Unlike existing methods, MIR3 does not require training with adversaries, yet is provably against allies executing worst-case actions. Theoretically, we formulate robust MARL as an inference problem, where the policy is trained in cooperative scenarios and implicitly maximize robust performance via off-policy evaluation. Under this formulation, we proof that minimizing mutual information serves as a lower bound for robustness. This objective can further be interpreted as suppressing spurious correlations through an information bottleneck, or as learning a robust action prior that encourage actions favored by the environment. In line of our theoretical findings, empirical results demonstrate that our MIR3 surpass baselines in robustness and training efficiency in StarCraft II and robot swarm control, and consistently exhibits superior robustness when deployed in real-world. Regarding limitations, our MIR3 is designed to be robust under many threat scenarios. When the task contains only one or very few threat scenarios, our method might not be as effective compared to methods that explicitly use max-min optimization against worst-case adversaries.

\bibliographystyle{unsrt}
\bibliography{neurips_2024}

\newpage

\appendix

{\LARGE\sc {Appendix for "Robust Multi-Agent Reinforcement Learning by Mutual Information Regularization"}\par}

\section{Proof for Proposition 1}
\label{appendix_proof}

The proof is constructed in three steps. First, we show the policy used by worst-case adversary and benign agents are inter-correlated. We then transform the benign policy in adversarial one. Second, we derive a lower bound for all attack trajectories and partitions. Third, we plug the lower bound to the cooperative case and get the final result.

\textbf{Step 1.} We first restate our objectives as follows:
\begin{equation}
\begin{split}
\label{eqn:perturb}
J(\pi) & = J^0(\pi) + \mathbb{E}_{\phi \in \Phi^\alpha} \left[ \min\limits_{\pi_{\alpha}} J^\phi(\pi) \right] \\
&= \sum_{t=0}^T \mathbb E_{\tau^0 \sim \hat{p}(\tau^0)} [r_t + \mathcal H(\pi(\mathbf a_t|\mathbf h_t))]  + \mathbb{E}_{\phi \in \Phi^\alpha} \big[ \min\limits_{\pi_{\alpha}} \sum_{t=0}^T  \mathbb E_{\tau^\phi \sim \hat{p}(\tau^\phi)} \left[r_t + \mathcal H(\pi(\mathbf a_t|\mathbf h_t))\right]\big].
\end{split}
\end{equation}
To proceed, the first step we take is to transform the policy in adversarial trajectories $\pi(\mathbf a_t|\mathbf h_t)$ into $\hat{\pi}(\hat{\mathbf a}_t|\mathbf h_t, \phi)$, such that the policy meets the trajectory probability with adversary. Recall in probabilistic reinforcement learning \cite{levine2018probinf}, the optimal policy is defined via soft Bellman backup:
\begin{equation}
\begin{split}
\label{eqn:perturb}
\pi(a_t|s_t) = \frac{1}{Z} \exp(Q(s_t, a_t) - V(s_t)),
\end{split}
\end{equation}
where Z is a normalizing constant.

This is extended to multi-agent reinforcement learning by marginalizing the actions of other agents \cite{wen2019pr2}. In our case, we further add current partition $\phi$ to the objective, which is written as:
\begin{equation}
\begin{split}
\label{eqn:perturb}
\pi(a_t^i|h_t^i) &= \frac{1}{Z} \exp(Q(s_t, a_t^i, a_t^{-i}, \phi) - Q(s_t, a_t^{-i}, \phi)) \\
&=\frac{1}{Z} \exp(Q(s_t, a_t^i, a_t^{-i}, \phi) - \log \int_{a^i_t} \exp( Q(s_t, a_t^i, a_t^{-i}, \phi) ) \mathrm d a^{i}_t ).
\end{split}
\end{equation}
Since the adversary is zero-sum, its objective is opposite to the objective of the defenders, which can be written as:
\begin{equation}
\begin{split}
\label{eqn:perturb}
\pi_\alpha(a_{t, \alpha}^i|h_t^i) &= \frac{1}{Z'} \exp( - Q(s_t, a_{t, \alpha}^i, a_t^{-i}, \phi) + Q(s_t, a_t^{-i}, \phi) ) \\
&= \frac{1}{Z'} \exp( - Q(s_t, a_{t, \alpha}^i, a_t^{-i}, \phi) + \log \int_{a^i_{t, \alpha}} \exp( Q(s_t, a_{t, \alpha}^i, a_t^{-i}, \phi) ) \mathrm d a_{t, \alpha}^{i} ).
\end{split}
\end{equation}
Next, we expand our objective in terms of history-action pairs, where history are added to meet the conditions of Dec-POMDP (\ie, policy always condition on current histories):
\begin{equation}
\begin{split}
\label{eqn:perturb}
J(\pi) &= -D_{KL}(\hat{p}(\tau^0)||p(\tau^0)) - \mathbb E_{\phi \in \Phi^\alpha} [ \min\limits_{\pi_{\alpha}} D_{KL}(\hat{p}(\tau^\phi)||p(\tau^\phi))] \\
&= \sum_{t=0}^T \mathbb E_{\mathbf h_t, \mathbf a_t \sim p(\tau^0)} [r_t - \log \pi(\mathbf a_t|\mathbf h_t)] + \mathbb{E}_{\phi \in \Phi^\alpha} \big[ \min\limits_{\pi_{\alpha}} \sum_{t=0}^T  \mathbb E_{\mathbf h_t, \hat{\mathbf a}_t \sim p(\tau^\phi)} \left[r_t - \log \pi(\mathbf a_t|\mathbf h_t)\right]\big]. \\
&= \sum_{t=0}^T \mathbb E_{\mathbf h_t, \mathbf a_t \sim p(\tau^0)} [r_t - \log \pi(\mathbf a_t|\mathbf h_t)] + \mathbb{E}_{\phi \in \Phi^\alpha} \Bigg[ \min\limits_{\pi_{\alpha}} \sum_{t=0}^T  \mathbb E_{\mathbf h_t, \hat{\mathbf a}_t \sim p(\tau^\phi)} \Bigg[r_t - \sum_{i=1}^{N} \log \pi^i(a_t^i|h_t^i)\Bigg]\Bigg].
\end{split}
\end{equation}
Here we cannot directly process the objective containing adversary since the trajectory is sampled using $\hat \pi(\hat{\mathbf a}_t|\mathbf h_t)$. Taking logarithm of current policy $\pi(a_t^i|h_t^i)$ and adversarial policy $\pi_\alpha(a_{t, \alpha}^i|h_t^i)$, we get:
\begin{equation}
\begin{split}
\label{eqn:perturb}
\log \pi(a_t^i|h_t^i) = -\log Z + Q(s_t, a_t^i, a_t^{-i}, \phi) - \log \int_{a^{i}_t} \exp( Q(s_t, a_t^i, a_t^{-i}, \phi) ) \mathrm d a^{i}_t.
\end{split}
\end{equation}
for defenders and 
\begin{equation}
\begin{split}
\label{eqn:perturb}
\log \pi_\alpha(a_{t, \alpha}^i|h_t^i) = -\log Z' - Q(s_t, a_{t, \alpha}^i, a_t^{-i}, \phi) + \log \int_{a^i_{t, \alpha}} \exp( Q(s_t, a_{t, \alpha}^i, a_t^{-i}, \phi) ) \mathrm d a_{t, \alpha}^i.
\end{split}
\end{equation}
for adversaries.

Thus, we have 
\begin{equation}
\begin{split}
\label{eqn:perturb}
\log \pi^\alpha(a_{t, \alpha}^i|h_t^i) = - \log \pi(a_t^i|h_t^i) + c,
\end{split}
\end{equation}
where $c = - \log Z + \log Z'$ is a constant. We ignore this in our subsequent derivations.

Plugging this into our objective, we get:
\begin{equation}
\begin{split}
\label{eqn:perturb}
J(\pi) &= -D_{KL}(\hat{p}(\tau^0)||p(\tau^0)) - \mathbb E_{\phi \in \Phi^\alpha} [ \min\limits_{\pi_{\alpha}} D_{KL}(\hat{p}(\tau^\phi)||p(\tau^\phi))] \\
&= \sum_{t=0}^T \mathbb E_{\mathbf h_t, \mathbf a_t \sim p(\tau^0)} [r_t - \log \pi(\mathbf a_t|\mathbf h_t)] + \mathbb{E}_{\phi \in \Phi^\alpha} \Bigg[  \min\limits_{\pi_{\alpha}} \sum_{t=0}^T  \mathbb E_{\mathbf h_t, \hat{\mathbf a}_t \sim p(\tau^\phi)} \left[r_t - \sum_{i=1}^{N} \log \pi^i(a_t^i|h_t^i)\right]\Bigg] \\
&= \sum_{t=0}^T \mathbb E_{\mathbf h_t, \hat{\mathbf a}_t \sim p(\tau^0)} [r_t - \log \pi(\mathbf a_t|\mathbf h_t)]  + \mathbb{E}_{\phi \in \Phi^\alpha} \Bigg[ \sum_{t=0}^T  \mathbb E_{\mathbf h_t, \hat{\mathbf a}_t \sim p(\tau^\phi)} \Bigg[r_t \\
& \ \ \ \ \ \ \ \ \ \ \ \ \ \ \ \ \ \ \ \ \ \ \ \ \ \ \ \ \ \ - \sum_{i \in \phi^i = 0} \log \pi(a_{t}|h_t^i) + \sum_{i \in \phi^i = 1} \log \pi_\alpha(a_{t, \alpha}^i|h_t^i, \phi) \Bigg]\Bigg].
\end{split}
\end{equation}
Now, for trajectories containing adversaries, the policy now correctly match the trajectories used for rollout.

\textbf{Step 2.} Next, we transform the objective containing adversarial rollouts into a regularization. In this step, we assume the history probability averaged across all partitions $\phi \in \Phi^\alpha$ follows a uniform distribution. While potentially strong, it serves as a reasonable prior considering all partitions where any agent could be adversarial when we have zero knowledge of the distribution under attack. Since we do not know the true history distribution under all adversarial attacks, a uniform distribution is at least guaranteed to cover all attack scenarios, such that defenders will not overlook certain conditions.

Starting from our previous objective, we get:
\begin{equation}
\begin{split}
\label{eqn:perturb}
J(\pi) &= -D_{KL}(\hat{p}(\tau^0)||p(\tau^0)) - \mathbb E_{\phi \in \Phi^\alpha} [ \min\limits_{\pi_{\alpha}} D_{KL}(\hat{p}(\tau^\phi)||p(\tau^\phi))] \\
&= \sum_{t=0}^T \mathbb E_{\mathbf h_t, \mathbf a_t \sim p(\tau^0)} [r_t - \log \pi(\mathbf a_t|\mathbf h_t)] + \mathbb{E}_{\phi \in \Phi^\alpha} \Bigg[ \sum_{t=0}^T  \mathbb E_{\mathbf h_t^, \hat{\mathbf a}_t \sim p(\tau^\phi)} \Bigg[r_t - \sum_{i \in \phi^i = 0} \log \pi(a_{t}|h_t^i) \\
& \ \ \ \ \ \ \ \ \ \ \ \ \ \ \ \ \ \ \ \ \ \ \ \ \ \ \ \ \ \  + \sum_{i \in \phi^i = 1} \log \pi_\alpha(a_{t, \alpha}^i|h_t^i, \phi) \Bigg]\Bigg] \\
&\geq \sum_{t=0}^T \mathbb E_{\mathbf h_t, \mathbf a_t \sim p(\tau^0)} [r_t - \log \pi(\mathbf a_t|\mathbf h_t)] + \mathbb{E}_{\phi \in \Phi^\alpha} \Bigg[ \sum_{t=0}^T  \mathbb E_{\mathbf h_t, \hat{\mathbf a}_t \sim p(\tau^\phi)} \Bigg[r_t + \sum_{i \in \phi^i = 0} \log \pi(a_{t}|h_t^i) \\
& \ \ \ \ \ \ \ \ \ \ \ \ \ \ \ \ \ \ \ \ \ \ \ \ \ \ \ \ \ \  + \sum_{i \in \phi^i = 1} \log \pi_\alpha(a_{t, \alpha}^i|h_t^i, \phi) \Bigg]\Bigg] \\
& = \sum_{t=0}^T \mathbb E_{\mathbf h_t, \mathbf a_t \sim p(\tau^0)} [r_t - \log \pi(\mathbf a_t|\mathbf h_t)] + \mathbb{E}_{\phi \in \Phi^\alpha} \Bigg[\sum_{t=0}^T \mathbb E_{\mathbf h_t, \hat{\mathbf a}_t \sim p(\tau^\phi)} [r_t + \log \hat{\pi}(\hat{\mathbf a}_t|\mathbf h_t) ] \Bigg]
\end{split}
\end{equation}
By plugging in the assumption that history is uniformly distributed (\ie, $p(\mathbf h_t) = \frac{1}{c}$), we get:
\begin{equation}
\begin{split}
\label{eqn:perturb}
J(\pi) & \geq \sum_{t=0}^T \mathbb E_{\mathbf h_t, \mathbf a_t \sim p(\tau^0)} [r_t - \log \pi(\mathbf a_t|\mathbf h_t)] + \mathbb{E}_{\phi \in \Phi^\alpha} \Bigg[ \sum_{t=0}^T \mathbb E_{\mathbf h_t, \hat{\mathbf a}_t \sim p(\tau^\phi)} [r_t + \log \hat{\pi}(\hat{\mathbf a}_t|\mathbf h_t) ] \Bigg] \\
& \geq  \sum_{t=0}^T \mathbb E_{\mathbf h_t, \mathbf a_t \sim p(\tau^0)} [r_t] + \mathbb E_{\mathbf h_t \sim p(\tau^0)} [\mathcal H(\mathbf a_t|\mathbf h_t)] + \mathbb{E}_{\phi \in \Phi^\alpha} \Bigg[ \sum_{t=0}^T \mathbb E_{\mathbf h_t, \hat{\mathbf a}_t \sim p(\tau^\phi)} [r_t] + \mathbb E_{\mathbf h_t \sim p(\tau^\phi)} [\log \pi(\mathbf a_t|\mathbf h_t)] \Bigg] \\
& \geq \sum_{t=0}^T \mathbb E_{\mathbf h_t, \mathbf a_t \sim p(\tau^0)} [r_t] + \mathbb E_{\mathbf h_t \sim p(\tau^0)} [\mathcal H(\mathbf a_t|\mathbf h_t)] +  \sum_{t=0}^T \mathbb E_{\mathbf h_t \sim p(\tau^\phi)} [\log \pi(\mathbf a_t|\mathbf h_t) + \log p(\mathbf h_t)] + c \\
& \geq \sum_{t=0}^T \mathbb E_{\mathbf h_t, \mathbf a_t \sim p(\tau^0)} [r_t] + \mathbb E_{\mathbf h_t \sim p(\tau^0)} [\mathcal H(\mathbf a_t|\mathbf h_t)] - \sum_{t=0}^T \mathcal H (\pi(\mathbf a_t, \mathbf h_t)) + c  \\
& \geq \sum_{t=0}^T \mathbb E_{\mathbf h_t, \mathbf a_t \sim p(\tau^0)} [r_t] + \mathbb E_{\mathbf h_t \sim p(\tau^0)} [\mathcal H(\mathbf a_t|\mathbf h_t)] - \sum_{t=0}^T \mathcal H (\pi(\mathbf a_t)) +\mathcal H( p(\mathbf h_t)) + c  \\
& \geq \sum_{t=0}^T \mathbb E_{\mathbf h_t, \mathbf a_t \sim p(\tau^0)} [r_t] + \mathbb E_{\mathbf h_t \sim p(\tau^0)} [\mathcal H(\mathbf a_t|\mathbf h_t)] - \mathcal H (\pi(\mathbf a_t)).
\end{split}
\end{equation}
\textbf{Step 3.} Finally, from information theory, we have:
\begin{equation}
\begin{split}
\label{eqn:perturb}
I(h_t; a_t) = \mathcal H(a_t) - \mathcal H(h_t, a_t),
\end{split}
\end{equation}
plugging in our derivations above, we get:
\begin{equation}
\begin{split}
\label{eqn:perturb}
J(\pi) & \geq \sum_{t=0}^T \mathbb E_{\mathbf h_t, \mathbf a_t \sim p(\tau^0)} [r_t] + \mathbb E_{\mathbf h_t \sim p(\tau^0)} [\mathcal H(\mathbf a_t|\mathbf h_t)] - \mathcal H (\pi(\mathbf a_t)) \\
&= \sum_{t=0}^T \mathbb E_{\mathbf h_t, \mathbf a_t \sim p(\tau^0)} [r_t] + \mathbb E_{\mathbf h_t \sim p(\tau^0)} [\mathcal H(\mathbf a_t|\mathbf h_t) - \mathcal H (\pi(\mathbf a_t))] \\
&= \sum_{t=0}^T \mathbb E_{\mathbf h_t, \mathbf a_t \sim p(\tau^0)} [r_t] + \mathbb E_{\mathbf h_t \sim p(\tau^0)} [- I(\mathbf h_t; \mathbf a_t)] \\
&= \sum_{t=0}^T \mathbb E_{\tau^0 \sim p(\tau^0)} [r_t - I(\mathbf h_t; \mathbf a_t)]
\end{split}
\end{equation}
This completes the proof. \qed

As a limitation, we acknowledge that the assumptions that the history distribution is uniform when averaged under all partitions $\phi \in \Phi$ may not hold for all environments, and the derived lower bound can be loose in some circumstances. However, our intuitive objective is effective as demonstrated by empirical results in both simulation environments and real world. We consider providing a more generalized proof or exploring alternative regularization that do not rely on these specific assumptions as future research.





\section{Algorithm for MIR3}
\label{appendix_algo}

Here we present our MIR3 defense algorithm. For all MARL algorithm, MIR3 first compute the mutual information between trajectories and actions, and subtract it from the reward received from environment. In this way, our MIR3 applies to all algorithms, with an example of using MADDPG backbone given in Algorithm. \ref{alg1}. For MIR3 defense with QMIX backbone, just change the way for parameter update from MADDPG to QMIX.

\begin{algorithm}[t]
\caption{MIR3 Defense with MADDPG backbone.}
\label{alg1}
\begin{algorithmic}[1]
\renewcommand{\algorithmicrequire}{\textbf{Input:}}
\renewcommand{\algorithmicensure}{\textbf{Output:}}
\REQUIRE Policy network of agents $\{\pi_1, \pi_2, ... \pi_N\}$, value function network $Q_i^\pi(s, a_1, ..., a_N)$, mutual information estimation network based on CLUB \cite{cheng2020club}: $CLUB(h_t^i, a_t^i)$, hyperparameter $\lambda$ for mutual information regularization.
\ENSURE Trained robust policy networks $\{\pi_1, \pi_2, ... \pi_N\}$.
\FOR{episode = 0, 1, 2, ... K}
\STATE Perform rollout using current policy, save trajectory in buffer $\mathcal D$.
\STATE Update $CLUB(h_t^i, a_t^i)$ using $\mathcal D$.
\STATE $I(\mathbf h_t, \mathbf a_t) \gets \sum_{i \in \mathcal N} CLUB(h_t^i, a_t^i)$.
\STATE $r_t^{MI} \gets r_t - \lambda \cdot I(\mathbf h_t, \mathbf a_t)$.
\STATE Update critic $\{Q_i\}$ of each agents using $r_t^{MI}$.
\STATE Update parameters of each agents using MADDPG. \texttt{// To implement MIR3 on other backbone, just change the way of parameter update.}
\ENDFOR
\end{algorithmic}
\end{algorithm}

\section{Implementation Details and Hyperparameters}
\label{appendix_details}

We implement our MIR3 and all baselines on the codebase of FACMAC \cite{fayad2021influencenips}, which empirically yields satisfying performance across many environments. For M3DDPG, the method is designed for robust continuous control, where actions are continuous and can be perturbed by a small value. To adapt M3DDPG to discrete control tasks, we add noise perturbation to the action probability of MADDPG, and send the perturbed action probability to the critic. We also find using large $\epsilon$ for M3DDPG will make the policy impossible to converge in fully cooperative settings: since M3DDPG add perturbations to the critic, a large perturbation renders the critic unable to fairly evaluate current status. As such, we select the largest $\epsilon$ which enables maximum robust performance while not cause cooperative training impossible in each setting.

The hyperparameters are listed as follows:

For \textbf{SMAC environment}, MIR3 and all baseline methods are implemented based on the same set of shared parameters, as listed in Table. \ref{params-smac-maddpg-all} and \ref{params-smac-qmix-all}. Parameters specific to MIR3 are listed in Table. \ref{params-smac-maddpg-MIR3} and \ref{params-smac-qmix-MIR3}. Note that for all experiments, the parameters of MIR3 do not change, except for $\lambda$. Empirically, we find $\lambda$ achieves best performance from 5e-5 to 5e-4.


\begin{table}[H]
\centering
\caption{Shared hyperparameters for SMAC on MADDPG backbone, used in MIR3 and all baselines.}
\label{params-smac-maddpg-all}
\begin{tabular}{cc|cc|cc}
\hline
Hyperparameter & Value & Hyperparameter & Value & Hyperparameter & Value\\ \hline
lr      &  5e-4 &  batch size   & 32 &  warmup steps      &    0 \\
  parallel envs &   1    &  buffer size  & 5000  &  $\tau$    &   0.05  \\
  gamma     &  0.99 &  evaluate episodes  & 32   &  train epochs   &  1  \\
 actor network &  RNN  &   exploratory noise &   0.1  &  num batches   &  1 \\
 hidden dim &  128  &   max grad norm  &   10   &   total timesteps &  5e6 \\
 hidden layer &  1   &   max episode len &  150  & M3DDPG $\epsilon$ & 0.003 \\
 activation   & ReLU   &  actor lr       &  =lr   &  ERNIE $K$    &  1   \\
 optimizer   & Adam   &  critic lr      &  =lr  &  ERNIE $\epsilon$ & 0.1 \\
  ROMAX $\kappa$  & 0.1   &  ROM-Q $P_{adv}$   & 0.3  &  & \\\hline
\end{tabular}
\vspace{-0.1in}
\end{table}

\begin{table}[H]
\centering
\caption{Shared hyperparameters for SMAC on QMIX backbone, used in MIR3 and all baselines.}
\label{params-smac-qmix-all}
\begin{tabular}{cc|cc|cc}
\hline
Hyperparameter & Value & Hyperparameter & Value & Hyperparameter & Value\\ \hline
lr      &  0.001 &  batch size   & 1000 &  warmup steps      &    5000 \\
  parallel envs &   1    &  buffer size  & 5000  &  $\tau$    &   0.005  \\
  gamma     &  0.99 &  evaluate episodes  & 20   &  train epochs   &  1  \\
 actor network &  MLP  &  $\epsilon$ start &   1.0  &  $\epsilon$ finish   &  0.05 \\
 $\epsilon$ anneal time &  100000  &   max grad norm  &   10   &   total timesteps &  4e6 \\
 hidden dim &  256   &   hidden layer &  1  & max episode len & 150 \\
 activation   & ReLU   &  mixing embedding dim &  32   &   hypernet layer    &  2  \\
 hypernet embedding   & 64   &  optimizer &  Adam  &  ROM-Q $P_{adv}$  & 0.3 \\
 ERNIE $K$  & 1 & ERNIE $\epsilon$ & 0.1 &  &  \\\hline
\end{tabular}
\vspace{-0.1in}
\end{table}


\begin{table}[H]
\centering
\setlength\tabcolsep{3pt}
\caption{Hyperparameters for MIR3 in SMAC environment on MADDPG backbone.}
\label{params-smac-maddpg-MIR3}
\begin{tabular}{cc | cc | cc}
\hline
Hyperparameter & Value & Hyperparameter & Value & Hyperparameter & Value\\ \hline
 $\lambda$   & 5e-4 &  MI buffer size & =buffer size &  MI train epochs & 1 \\   
 MI lr     & =lr    &  MI hidden dim & =hidden dim  &      &    \\ \hline
\end{tabular}
\end{table}

\begin{table}[H]
\centering
\setlength\tabcolsep{3pt}
\caption{Hyperparameters for MIR3 in SMAC environment on QMIX backbone.}
\label{params-smac-qmix-MIR3}
\begin{tabular}{cc | cc | cc}
\hline
Hyperparameter & Value & Hyperparameter & Value & Hyperparameter & Value\\ \hline
 $\lambda$   & 1e-4 &  MI Buffer size & =buffer size &  MI train epochs & 1 \\   
 MI lr     & =lr    &  MI hidden dim & =hidden dim  &      &    \\ \hline
\end{tabular}
\end{table}

For \textbf{rendezvous}, all methods are implemented based on the same set of shared parameters, which mainly follows MAMujoco, as listed in Table. \ref{params-rendezvous-maddpg-all}. For MIR3, its parameters are listed in Table. \ref{params-rendezvous-maddpg-MIR3}.

\begin{table}[H]
\centering
\caption{Shared hyperparameters for rendezvous on MADDPG backbone, used in MIR3 and all baselines.}
\label{params-rendezvous-maddpg-all}
\begin{tabular}{cc|cc|cc}
\hline
Hyperparameter & Value & Hyperparameter & Value & Hyperparameter & Value\\ \hline
lr      &  1e-3 &  batch size   & 8 &  warmup steps      &    0 \\
  parallel envs &   1    &  buffer size  & 5000  &  $\tau$    &   0.01  \\
  gamma     &  0.99 &  evaluate episodes  & 32   &  train epochs   &  1  \\
 actor network &  MLP  &   exploratory noise &   0.1  &  num batches   &  1 \\
 hidden dim &  256  &   max grad norm  &   0.5   &   total timesteps &  1e7 \\
 hidden layer &  1   &   max episode len &  200  & M3DDPG $\epsilon$ & 0.001 \\
 activation   & ReLU   &  actor lr       &  =lr   & ERNIE $K$ & 1 \\
 optimizer   & Adam   &  critic lr      &  =lr  & ERNIE $\epsilon$ & 0.1 \\
  ROMAX $\kappa$   & 0.01   &  ROM-Q $P_{adv}$ &  0.1  & ROM-Q $\epsilon$ & 1 \\\hline
\end{tabular}
\vspace{-0.1in}
\end{table}

\begin{table}[H]
\centering
\setlength\tabcolsep{3pt}
\caption{Hyperparameters for MIR3 in rendezvous environment on MADDPG backbone.}
\label{params-rendezvous-maddpg-MIR3}
\begin{tabular}{cc | cc | cc}
\hline
Hyperparameter & Value & Hyperparameter & Value & Hyperparameter & Value\\ \hline
 $\lambda$   & 5e-5 &  MI Buffer size & =buffer size &  MI train epochs & 1 \\   
 MI lr     & =lr    &  MI hidden dim & =hidden dim  &      &    \\ \hline
\end{tabular}
\end{table}




\section{Results With Many Adversaries}
\label{appendix_twoadv}

\begin{figure*}[h]
\centering
\includegraphics[scale=0.22]{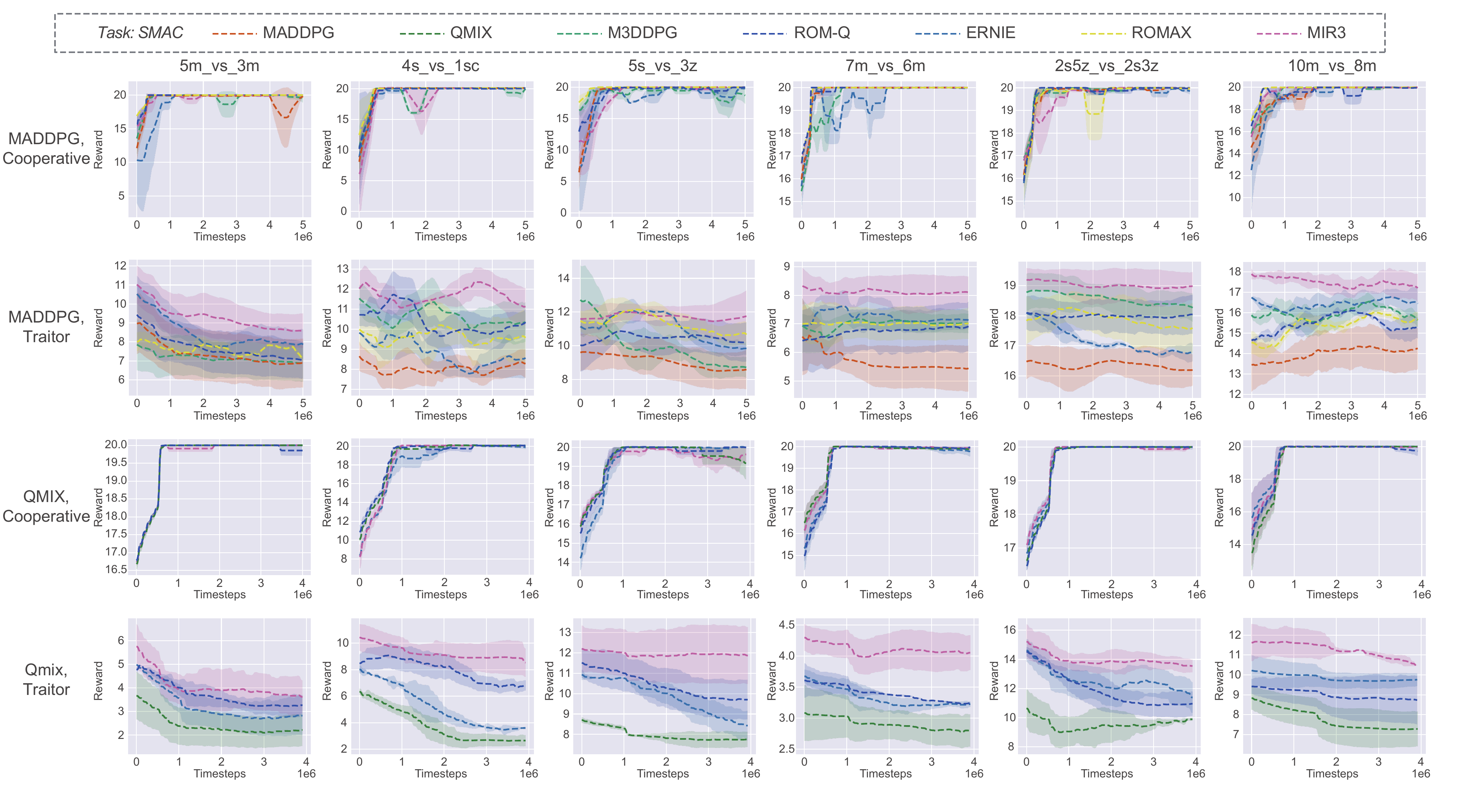}
\caption{Results of two adversaries on six SMAC tasks, evaluated on MADDPG and QMIX backbones. The robustness of our MIR3 consistently outperform baseline results.}
\label{twoadv_supp}
\end{figure*}

Here we present the additional results of having two adversaries in the game, evaluated in the same six tasks in SMAC environment. Note that \emph{5m vs 3m} is reported in main paper. As shown in Fig. \ref{twoadv_supp}, in line with the results reported in our main paper, our MIR3 consistently outperform all baselines in robustness, while not compromising cooperative performance.

\section{Learned Robust Behaviors in SMAC \emph{9m vs 8m}}
\label{appendix_smac_video}

\begin{figure*}[h]
\centering
\includegraphics[scale=0.18]{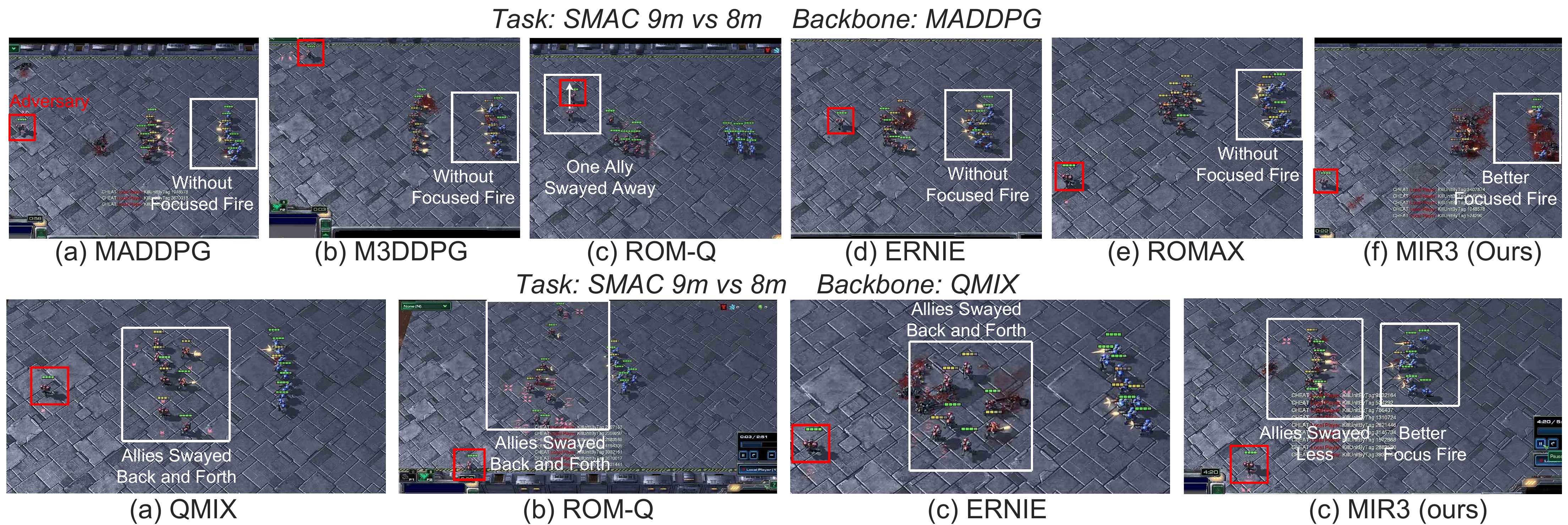}
\caption{Agent behaviors under attack in task \emph{9m vs 8m}, adversary denoted by red square. Under MADDPG backbone, baselines are either swayed by adversaries or lack cooperation on focused fire. Under QMIX backbone, baselines agents are frequently swayed back and forth. In contrast, our MIR3 is less swayed by adversary.}
\label{9m_vs_8m}
\end{figure*}

The behavior of algorithms in task \emph{9m vs 8m} is similar to \emph{4m vs 3m}, despite all agents generally perform worst, showing there is still a long path towards realizing robust MARL algorithm. Specifically, as illustrated in Fig. \ref{9m_vs_8m}, under the MADDPG backbone, baseline algorithms generally do not focused fire on enemies. In ROM-Q, one benign agent was even swayed by allies. Via reducing mutual information, our MIR3 MIR3 ensures that agents are not swayed and maximally maintain focused fire behavior under attack. However, with many agents, even our MIR3 can fail to coordinate occasionally, not attacking the same enemy at the same time.

Under the QMIX backbone, almost no methods are able to cooperate. Agents trained by QMIX, ROM-Q and ERNIE are swayed back and forth by adversaries without attack, thus easily being eliminated by enemy. In contrast, our MIR3 agents are still swayed, but much less than agents trained by baseline algorithm, thus eventually win the game.

\section{Learned Robust Behaviors in Robot Swarm Control}
\label{appendix_video}

\begin{figure*}[!htb]
\centering
\includegraphics[scale=0.35]{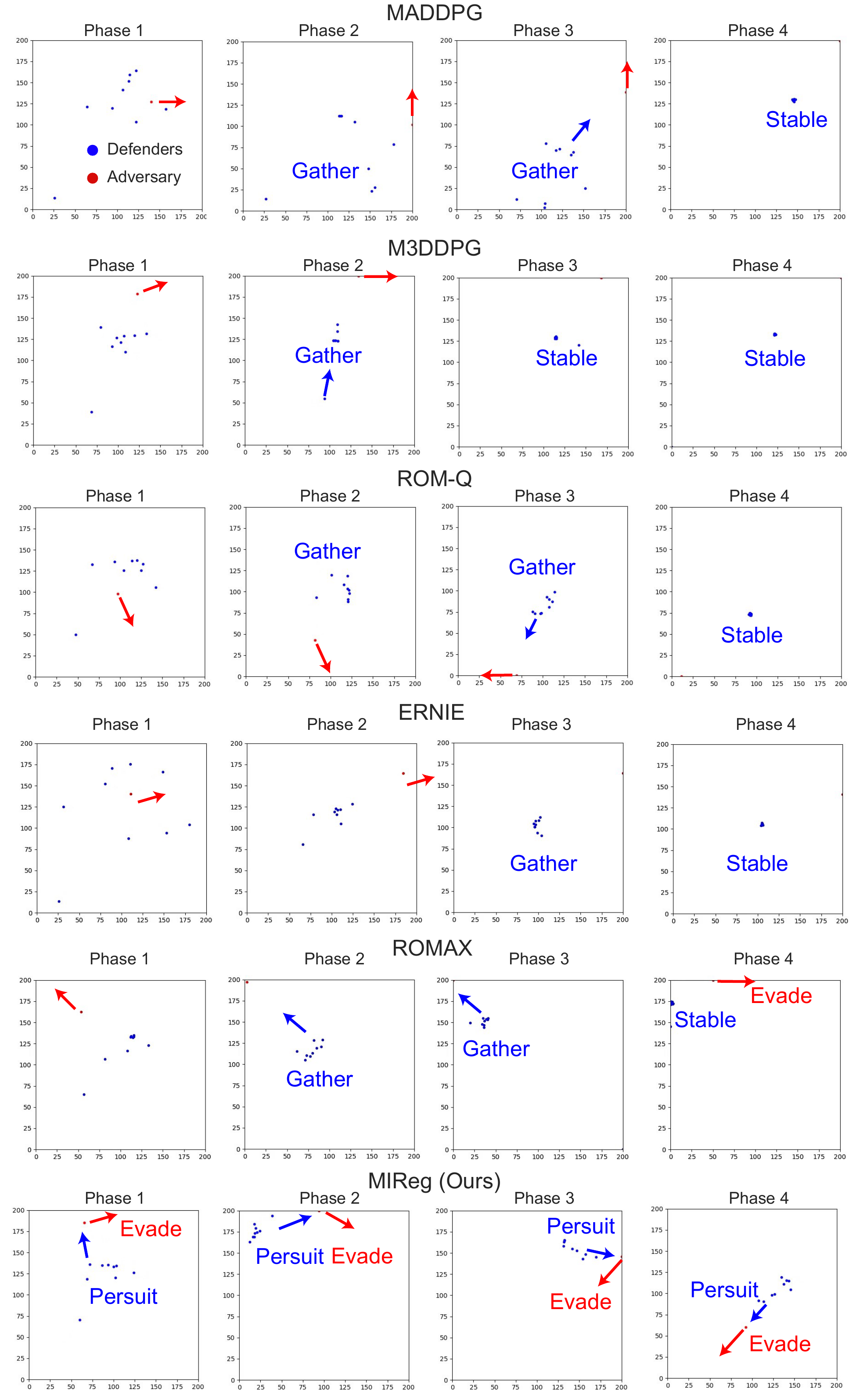}
\caption{Illustration of the robust behaviors in rendezvous environment. MADDPG can be fooled by the adversary and gets dispersed. M3DDPG, ROM-Q and ERNIE are less swayed, but gets together too fast and too close, thus unable to merge with adversaries together. ROMAX agents learns to gather and chase the adversary. However, eventually the agent remains stable which makes the adversary evade easily. In stark contrast, while learned in the environment to gather together, our MIR3 evolves distinctive pursuit-evade behavior in the presence of adversary.}
\label{rende}
\end{figure*}

\textbf{Rendezvous.} We report the learned robust behavior of rendezvous, where our MIR3 displays distinctive and superior behavior. As shown in Fig. \ref{rende}, MADDPG can be easily fooled by adversaries, such that adversary first sways the majority away from the agents that are not gathered, resulting in agents gathering together slower. Besides, during training, agents gets fixed after gathering together. As a consequence, MADDPG agents learns a spurious correlation that agents should always gather together tightly first, and wait others to join in the group. However, since the adversary will never came, the agents with spurious correlations have to wait forever and never being able to get together. For M3DDPG, ROM-Q and ERNIE, agents are not swayed to have longer gather time. However, agents still falls short on spurious correlations and can only move to the adversaries jointly in a very slow speed, and eventually stay still far away from the adversary. ROMAX agents are able to slowly gather and collaborate pursuit the adversary, yet it finally stays stable and the adversary are able to get away quickly.

In stark contrast, despite never encountering the adversary, by minimizing mutual information, our MIR3 implicitly suppresses the spurious correlations of agents, and emerges the behavior of pursuit-evade, despite never seeing the adversary. This is a clear evidence that suppressing mutual information enhances resilient and adaptive behavior of agents by countering spurious correlations.

Note that the behaviors in simulation is different comparing with behaviors in real world robot control. In real world, robots will collide, have different underlying physics, including different friction and robot dynamics \etc \ comparing with simulation, and can receive inaccurate sensing signal or take inaccurate actions. Thus, we do not see such pursuit-evade behaviors in our real world robots experiments. Instead, in real world, our agents gather together without being swayed, which also secures the highest reward in real world.

\section{Discussions on Considering Worst-Case Scenario}
\label{appendix_romq}

\begin{figure*}[h]
\centering
\includegraphics[scale=0.3]{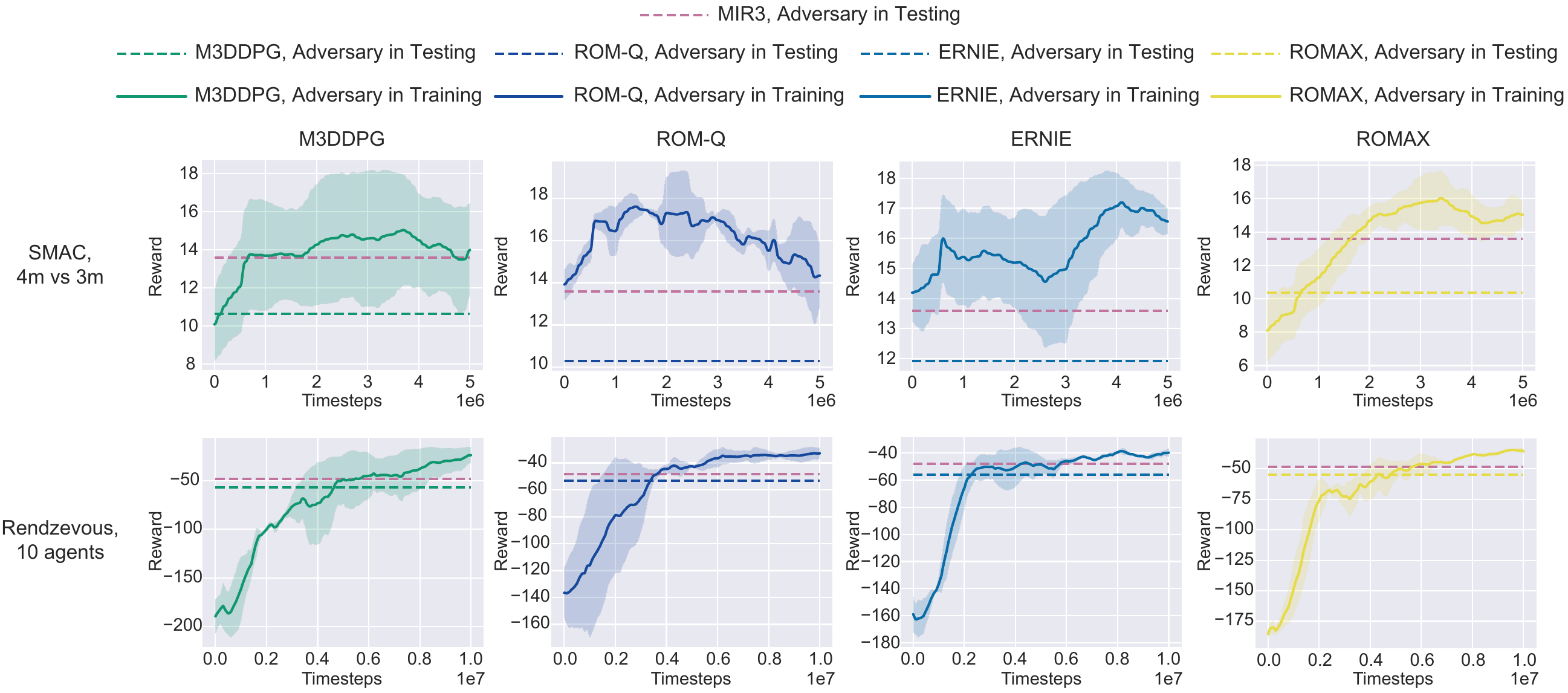}
\caption{Discussions on robust baselines considering threat scenarios with approximated adversaries during training and against worst-case adversaries during testing. The adversaries used by baselines in training are either inaccurate or insufficiently strong, resulting in high defense capability in training, but not in testing against worst-case adversaries.}
\label{loadckpt}
\end{figure*}

We add an additional discussion on the ineffectiveness of robust baselines considering worst-case scenario. Specifically, we compared the results of robust baselines when encountering the adversaries during training time and their effectiveness under the worst-case adversary at test time. As shown in Fig. \ref{loadckpt}, robust baselines perform well during training, yet failed when encountering worst-case adversaries. To understand this, for M3DDPG, ERNIE and ROMAX that consider all agents as potential adversaries, their defense slightly deviate the original policy, resulting a defense that are either too conservative when the deviation is large, and defense that are too weak when the deviation is small. As a result, the defenses generally falls short when encountering our strong worst-case adversaries.

For ROM-Q that considers each threat scenario, we hypothesis the insufficient defense capability of ROM-Q against worst-case adversaries can be attributed to the insufficient approximation of worst-case adversaries during training. Indeed, ROM-Q achieves higher reward in training time, showing the policy is not sufficiently equilibrated and the optimal worst-case adversarial policy has not been found. As a consequence, the trained policy using ROM-Q is only effective against weak adversaries, and cannot withstand the worst-case adversary during testing. In contrast, our MIR3 do not use adversary for training, yet is still more robust under worst-case adversaries.

\newpage
\section*{NeurIPS Paper Checklist}

\begin{enumerate}

\item {\bf Claims}
    \item[] Question: Do the main claims made in the abstract and introduction accurately reflect the paper's contributions and scope?
    \item[] Answer: \answerYes{}  
    \item[] Justification: Yes, we have reflected the contributions and scope of this paper.

\item {\bf Limitations}
    \item[] Question: Does the paper discuss the limitations of the work performed by the authors?
    \item[] Answer: \answerYes{} 
    \item[] Justification: We have discussed limitations in Conclusions section..

\item {\bf Theory Assumptions and Proofs}
    \item[] Question: For each theoretical result, does the paper provide the full set of assumptions and a complete (and correct) proof?
    \item[] Answer: \answerYes{} 
    \item[] Justification: We have followed the suggestions in guidelines.

    \item {\bf Experimental Result Reproducibility}
    \item[] Question: Does the paper fully disclose all the information needed to reproduce the main experimental results of the paper to the extent that it affects the main claims and/or conclusions of the paper (regardless of whether the code and data are provided or not)?
    \item[] Answer: \answerYes{} 
    \item[] Justification: We have released the code in supplementary materials. We have provided all the hyperparameters needed.

\item {\bf Open access to data and code}
    \item[] Question: Does the paper provide open access to the data and code, with sufficient instructions to faithfully reproduce the main experimental results, as described in supplemental material?
    \item[] Answer: \answerYes{} 
    \item[] Justification: We have provided the code in supplementary material, which we will make it open source after this paper is accepted.

\item {\bf Experimental Setting/Details}
    \item[] Question: Does the paper specify all the training and test details (e.g., data splits, hyperparameters, how they were chosen, type of optimizer, etc.) necessary to understand the results?
    \item[] Answer: \answerYes{} 
    \item[] Justification: All details are provided in appendix. Also check the code.

\item {\bf Experiment Statistical Significance}
    \item[] Question: Does the paper report error bars suitably and correctly defined or other appropriate information about the statistical significance of the experiments?
    \item[] Answer: \answerYes{} 
    \item[] Justification: We have provided error bars.

\item {\bf Experiments Compute Resources}
    \item[] Question: For each experiment, does the paper provide sufficient information on the computer resources (type of compute workers, memory, time of execution) needed to reproduce the experiments?
    \item[] Answer: \answerYes{} 
    \item[] Justification: We have provided all CPU and GPU informations, as stated in our experiments discussing training efficiency.
    
\item {\bf Code Of Ethics}
    \item[] Question: Does the research conducted in the paper conform, in every respect, with the NeurIPS Code of Ethics \url{https://neurips.cc/public/EthicsGuidelines}?
    \item[] Answer: \answerYes{} 
    \item[] Justification: We have obeyed the NeurIPS Code of Ethics.

\item {\bf Broader Impacts}
    \item[] Question: Does the paper discuss both potential positive societal impacts and negative societal impacts of the work performed?
    \item[] Answer: \answerNo{} 
    \item[] Justification: Our paper works on the robustness of multi-agent reinforcement learning, which will not result in negative societal impacts. The rest of the paper discuss how to enhance robustness, which will bring positive social impact and we believe it is not needed to discuss it separately.
    
\item {\bf Safeguards}
    \item[] Question: Does the paper describe safeguards that have been put in place for responsible release of data or models that have a high risk for misuse (e.g., pretrained language models, image generators, or scraped datasets)?
    \item[] Answer: \answerNo{} 
    \item[] Justification: Our work is to enhance the robustness of MARL. We believe our algorithm itself serves as a safeguard.

\item {\bf Licenses for existing assets}
    \item[] Question: Are the creators or original owners of assets (e.g., code, data, models), used in the paper, properly credited and are the license and terms of use explicitly mentioned and properly respected?
    \item[] Answer: \answerYes{} 
    \item[] Justification: We have cited the papers where the assets were used. We follow their instructions in Github, with Apache-2.0 license.

\item {\bf New Assets}
    \item[] Question: Are new assets introduced in the paper well documented and is the documentation provided alongside the assets?
    \item[] Answer: \answerNo{} 
    \item[] Justification: We do not release new assets.

\item {\bf Crowdsourcing and Research with Human Subjects}
    \item[] Question: For crowdsourcing experiments and research with human subjects, does the paper include the full text of instructions given to participants and screenshots, if applicable, as well as details about compensation (if any)? 
    \item[] Answer: \answerNA{} 
    \item[] Justification: Our paper does not involve crowdsourcing nor research with human subjects.

\item {\bf Institutional Review Board (IRB) Approvals or Equivalent for Research with Human Subjects}
    \item[] Question: Does the paper describe potential risks incurred by study participants, whether such risks were disclosed to the subjects, and whether Institutional Review Board (IRB) approvals (or an equivalent approval/review based on the requirements of your country or institution) were obtained?
    \item[] Answer: \answerNA{} 
    \item[] Justification: Our paper does not involve crowdsourcing nor research with human subjects.

\end{enumerate}

\end{document}